\documentclass[letterpaper]{article} % DO NOT CHANGE THIS
\usepackage{aaai2026}  % DO NOT CHANGE THIS
\usepackage{times}  % DO NOT CHANGE THIS
\usepackage{helvet}  % DO NOT CHANGE THIS
\usepackage{courier}  % DO NOT CHANGE THIS
\usepackage[hyphens]{url}  % DO NOT CHANGE THIS
\usepackage{graphicx} % DO NOT CHANGE THIS
\usepackage{natbib}  % DO NOT CHANGE THIS AND DO NOT ADD ANY OPTIONS TO IT
\usepackage{caption} % DO NOT CHANGE THIS AND DO NOT ADD ANY OPTIONS TO IT
\usepackage{algorithm}
\usepackage{algorithmic}

\usepackage{multirow}
\usepackage{amsthm, amsmath, amssymb}
\usepackage{times}
\usepackage{helvet}
\usepackage{courier}
\usepackage{xcolor}

\usepackage{newfloat}
\usepackage{listings}
\DeclareCaptionStyle{ruled}{labelfont=normalfont,labelsep=colon,strut=off} % DO NOT CHANGE THIS
\floatstyle{ruled}
\newfloat{listing}{tb}{lst}{}
\floatname{listing}{Listing}

\title{Skeletons Speak Louder than Text: A Motion-Aware Pretraining Paradigm for Video-Based Person Re-Identification}

\author{
    Rifen Lin\textsuperscript{\rm 1}, 
    Alex Jinpeng Wang\textsuperscript{\rm 1}, 
    Jiawei Mo\textsuperscript{\rm 1}, 
    Min Li\textsuperscript{\rm 1}\thanks{Corresponding author}
}
\affiliations{
    \textsuperscript{\rm 1}School of Computer Science and Engineering, Central South University\\
    rifen\_lin@csu.edu.cn, limin@mail.csu.edu.cn
}

\usepackage{bibentry}
\begin{document}

\maketitle

\begin{abstract}
Multimodal pretraining has revolutionized visual understanding, but its impact on video-based person re-identification (ReID) remains underexplored. 
Existing approaches often rely on video-text pairs, yet suffer from two fundamental limitations: (1) lack of genuine multimodal pretraining, and (2) text poorly captures fine-grained temporal motion—an essential cue for distinguishing identities in video.
In this work, we \textbf{take a bold departure from text-based paradigms by introducing the first skeleton-driven pretraining framework for ReID}.
To achieve this, we propose Contrastive Skeleton-Image Pretraining for ReID (CSIP-ReID), a novel two-stage method that leverages skeleton sequences as a spatiotemporally informative modality aligned with video frames.
In the first stage, we employ contrastive learning to align skeleton and visual features at sequence level.
In the second stage, we introduce a dynamic Prototype Fusion Updater (PFU) to refine multimodal identity prototypes, fusing motion and appearance cues. 
Moreover, we propose a Skeleton Guided Temporal Modeling (SGTM) module that distills temporal cues from skeleton data and integrates them into visual features.
Extensive experiments demonstrate that CSIP-ReID achieves new state-of-the-art results on standard video ReID benchmarks (MARS, LS-VID, iLIDS-VID). 
Moreover, it exhibits strong generalization to skeleton-only ReID tasks (BIWI, IAS), significantly outperforming previous methods.
CSIP-ReID \textbf{pioneers an annotation-free and motion-aware pretraining paradigm for ReID, opening a new frontier in multimodal representation learning}.
\end{abstract}

% % Uncomment the following to link to your code, datasets, an extended version or similar.
% % You must keep this block between (not within) the abstract and the main body of the paper.
% % \begin{links}
% %     \link{Code}{https://aaai.org/example/code}
% %     \link{Datasets}{https://aaai.org/example/datasets}
% %     \link{Extended version}{https://aaai.org/example/extended-version}
% % \end{links}

\section{Introduction}

\begin{figure}[h]
	\centering
	\includegraphics[width=0.95\linewidth]{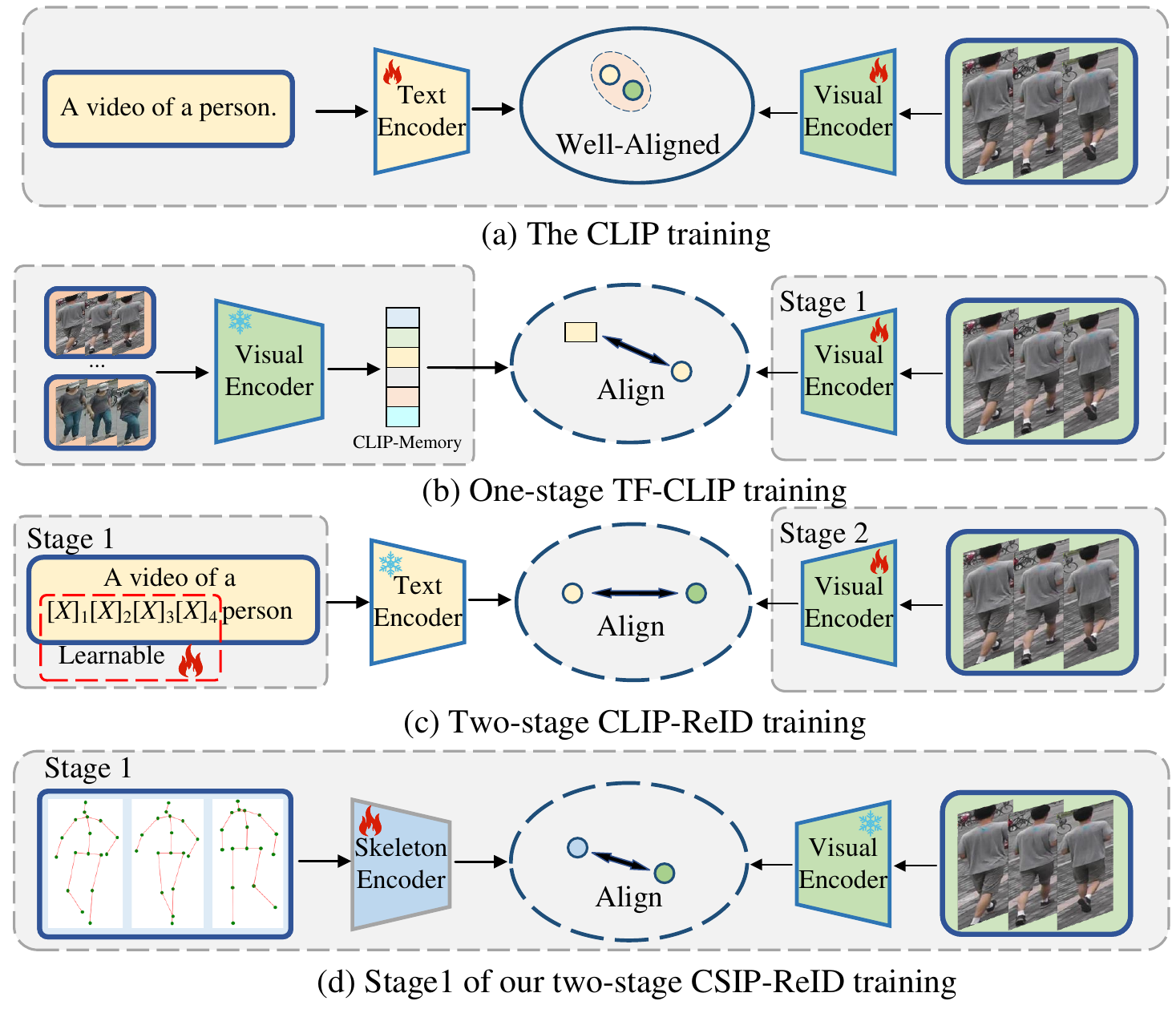}
	\caption{ We propose the first contrastive skeleton-image pretraining for ReID.
    Comparison of CLIP-style learning frameworks: 
    (a) CLIP training. (b) one-stage TF-CLIP training. (c) Two-stage CLIP-ReID training. (d) Contrastive learning in stage 1 of our two-stage CSIP-ReID training.
	}
	\label{fig:intro}
\end{figure}

Pretraining has profoundly transformed various areas of computer vision, from image classification ~\cite{chen2020simple} to multimodal understanding ~\cite{li2021align}, by learning transferable and robust representations from large-scale unlabeled data.
In particular, contrastive pretraining frameworks like CLIP~\cite{radford2021learning} have demonstrated remarkable generalization by aligning visual and textual modalities, enabling zero-shot and few-shot capabilities across downstream tasks ~\cite{xu2021videoclip,zhou2022learning}.
Although progress has been made in vision pretraining, video-based person re-identification, which matches individuals across non-overlapping cameras, remains underexplored in terms of cross-modal and pretraining approaches.

Some recent studies have attempted to bring CLIP-style frameworks into the ReID domain~\cite{li2023clip, tfclip_2024,li2025video}.
However, these methods \textbf{\textit{do not perform genuine multimodal pretraining using ReID datasets}}.
As shown in Fig.~\ref{fig:intro}, they typically reuse the pretrained CLIP visual encoder or text encoder, which is trained on generic image-text pairs that describe objects or scenes rather than person identities.
Such text \textbf{\textit{lacks the identity-level semantics necessary for fine-grained discrimination}}, making the learned modality alignment poorly suited for ReID.
Furthermore, these methods neglect the temporal dimension entirely, as \textbf{\textit{their training process does not incorporate motion cues or sequential modeling}}, which are crucial in video-based re-identification.

To overcome the limitations of prior works, we take a fundamentally different approach.
We introduce \textbf{\textit{the first framework that conducts genuine multimodal pretraining directly on ReID datasets}}, rather than relying on frozen encoders or handcrafted templates derived from unrelated domains.
Instead of using text as the second modality, we leverage skeleton sequences—a rich, structured, and annotation-free source of motion information that is naturally aligned with visual inputs in video.

Building on this foundation, we present \textbf{\textit{Contrastive Skeleton-Image Pretraining for ReID (CSIP-ReID)}}, a two-stage framework that learns joint representations from paired skeleton–image sequences.
Skeleton data offers several key advantages over text.
It encodes fine-grained motion patterns that are highly discriminative for person identification, remains robust under appearance or viewpoint variations, and can be efficiently extracted from videos using modern pose estimation models~\cite{goel2023humans, shen2024world}.
By replacing noisy or generic textual descriptions with expressive motion features, CSIP-ReID establishes a scalable and identity-aware pretraining paradigm tailored specifically for ReID.

After pretraining, we adopt prototype learning for identity supervision, as it effectively aggregates intra-class diversity, including variations in viewpoint and motion, and improves robustness against noisy samples.
Specifically, we propose a \textbf{\textit{Prototype Fusion Updater (PFU)}}, which integrates aligned appearance-rich visual features and motion-capturing skeleton features to generate more discriminative and robust prototypes than previous methods ~\cite{tfclip_2024}. 
This is achieved by:
(1) Discarding empty frames via skeleton detection;
(2) Leveraging background-free skeleton representations to minimize redundancy;
(3) Fusing complementary appearance and motion information.

Since the visual encoder lacks spatiotemporal modeling, temporal information across frames is often ignored, causing the task to degenerate into image-based ReID.
To address this, we introduce a \textbf{\textit{Skeleton Guided Temporal Modeling (SGTM)}} module, which captures temporal dynamics through three components: Message Token Encoding (MTE), Auxiliary Temporal Distillation (ATD), and Temporal Aggregation (TA).
SGTM distills the strong temporal modeling capability of skeleton as guidance to enhance temporal representation following Learning Using Privileged Information (LUPI) ~\cite{vapnik2009new} paradigm.

Our main contributions can be summarized as follows:
\begin{itemize}
    \item We propose \textbf{\textit{CSIP-ReID}}, the first skeleton-driven pretraining framework for video-based ReID that learns from paired skeleton–image sequences. Unlike prior works that reuse CLIP encoders, our method performs genuine multimodal pretraining on ReID data, \textbf{\textit{establishing a new paradigm beyond text-based approaches}}.
    
    \item We introduce skeletons as a scalable, annotation-free alternative to text for contrastive pretraining. Skeletons are \textbf{\textit{inherently spatiotemporal and identity-discriminative}}, making them well-suited for motion-aware representation learning in video ReID.

    \item We design a Prototype Fusion Updater (PFU) using prototype learning to guide visual encoder finetuning and a Skeleton Guided Temporal Modeling (SGTM) module to distill temporal cues from skeletons.
    
    \item Our method achieves state-of-the-art performance on both video-based and skeleton-based ReID benchmarks, showcasing its effectiveness and generalization.
\end{itemize}

\section{Related Work}

\subsection{Video-based Person Re-Identification}

Video-based person re-identification aims to extract informative spatial-temporal cues from video sequences to learn robust identity representations. 
Early works employ CNNs \cite{he2021dense,liu2023video} or vision Transformers \cite{wu2022cavit,wang2025learning} to capture spatial features. 
Recently, TF-CLIP \cite{tfclip_2024} introduced a CLIP-style approach that replaces the text encoder with a visual memory module.
However, it remains unimodal and lacks genuine cross-modal contrastive pretraining.
In contrast, our CSIP-ReID performs genuine contrastive pretraining on paired skeleton–image sequences.

For temporal information extraction, existing methods adopt RNNs~\cite{dai2018Video}, 3D CNNs~\cite{gu2020appearance}, temporal pooling~\cite{wu2018exploit}, attention mechanisms~\cite{liu2023video} or temporal diffusion~\cite{tfclip_2024} to capture cross-frame temporal information. 
Unlike existing methods, we propose a Skeleton Guided Temporal Modeling (SGTM) module that uses skeletons as privileged information to guide temporal feature learning, following the LUPI paradigm~\cite{vapnik2009new}.

\subsection{Visual Skeleton Learning}

Recent studies have demonstrated the effectiveness of combining skeleton and visual modalities across various tasks.
Shao \emph{et al.} ~\cite{shao2021multi} integrate silhouette image and skeleton features through multimodal fusion.
Jiang \emph{et al.} ~\cite{jiang2024skeleton} enhance visible-infrared person ReID by guiding visual feature refinement and body-part fusion with skeleton graph modeling.
Liu \emph{et al.} ~\cite{liu2024multi} align visual features and skeleton features via contrastive learning.
Lu \emph{et al.} ~\cite{lu2024cross} transfer high-quality features from X-CLIP to skeleton encoder.
3DAPRL ~\cite{jing20253d} leverages 3D pedestrian representations, which are highly similar to skeleton, and introduce shape-aware spatio-temporal modeling to enhance video-based person ReID.

Despite these advances, existing approaches incorporate skeleton by adding separate streams or modules, inevitably increasing model complexity and computational cost. 
In the era of large-scale models, boosting performance without significantly increasing model size or runtime is essential, techniques such as pretraining ~\cite{chen2023vlp} and knowledge distillation ~\cite{xu2024survey} offer promising solutions.
Inspired by this, CSIP-ReID adopts a skeleton-image contrastive pretraining strategy and distills skeleton information employing prototype learning and LUPI paradigm.

\begin{figure*}[t]
	\centering
	\includegraphics[width=1.0\textwidth]{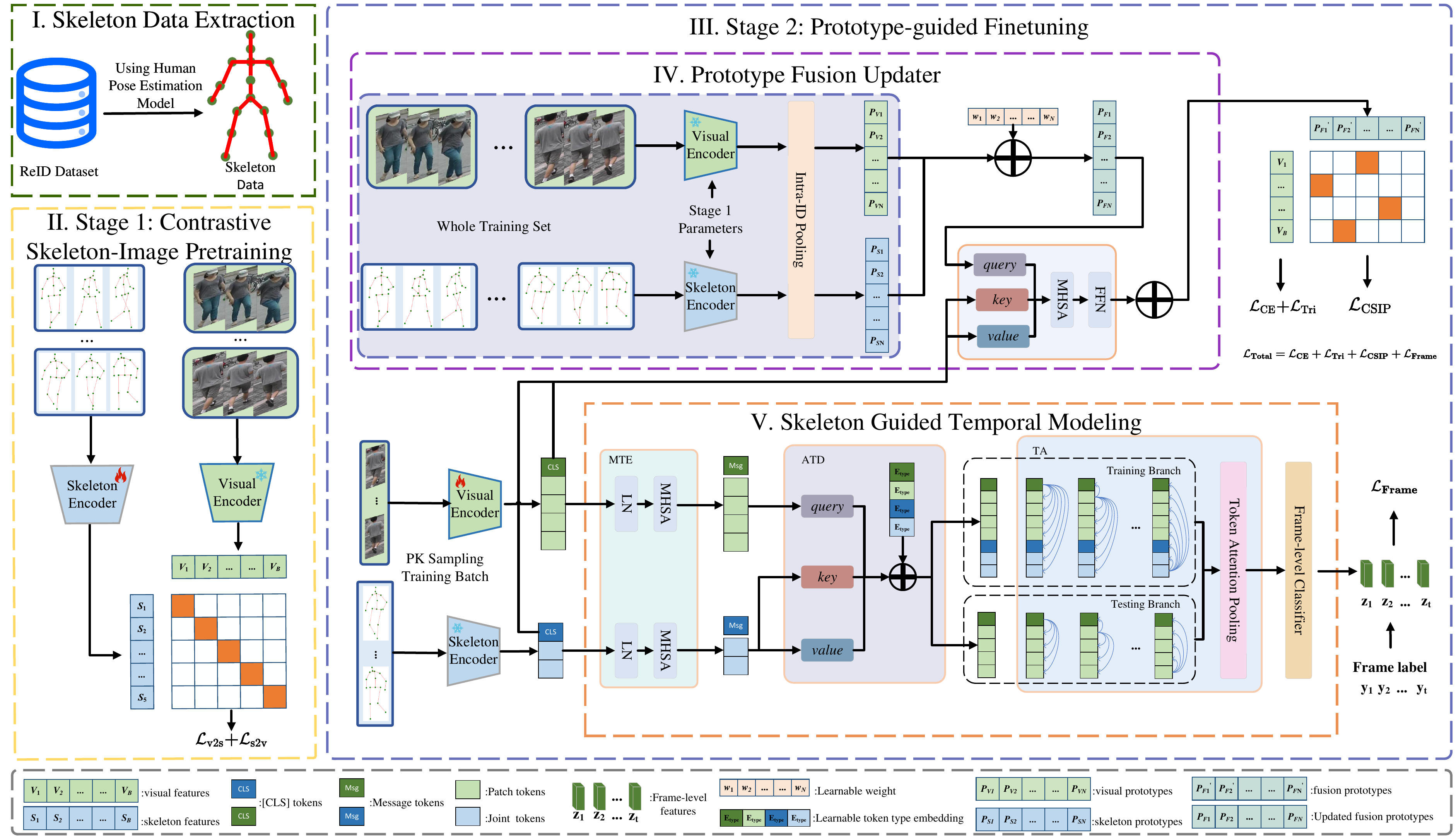}
	\caption{Illustration of the proposed CSIP-ReID framework. (I) Extract skeleton data using human pose estimation model. (II) Stage 1: Contrasitive Skeleton-Image Pretraining. (III) Stage 2: Prototype-guided Finetuning, consisting of Prototype Fusion Updater (PFU) and Skeleton Guided Temporal Modeling (SGTM). (IV) Prototype Fusion Updater (PFU) computes and fuses modality-specific prototypes, dynamically updating them with batch visual-skeleton features. (V) Skeleton Guided Temporal Modeling (SGTM) uses MTE to generate message tokens, employs ATD to distill skeleton temporal cues into visual features, and applies TA to aggregate these cues across tokens for frame-level representation.}
	\label{fig:overview}
\end{figure*}

\section{Method}

In this section, we present CSIP-ReID, a two-stage framework illustrated in Fig.~\ref{fig:overview}. We describe feature extraction, introduce Stage 1 for contrastive skeleton–image pretraining, and Stage 2 for prototype-guided finetuning with PFU and SGTM modules, followed by the overall training procedure.

\subsection{Feature Extraction with Encoders}  

CSIP-ReID consists of a \textbf{\textit{visual encoder $\mathcal{V}(\cdot)$}} and a \textbf{\textit{skeleton encoder $\mathcal{S}(\cdot)$}}.
We adopt \textbf{\textit{Vision Transformer (ViT)}} as the visual encoder for its strong spatial modeling capability and proven effectiveness in person re-identification ~\cite{he2021transreid}.
Meanwhile, \textbf{\textit{Skeleton Graph Transformer (SGT)}} is employed as the skeleton encoder to capture spatio-temporal patterns from joint graphs, owing to its strong performance in skeleton-based ReID ~\cite{rao2023transg}.

For the visual modality, an input sequence $\mathbf{V}=\{ \mathbf{V}_t \}_{t=1}^{T}$ with frames $\mathbf{V}_t \in \mathbb{R}^{H \times W \times 3}$ is encoded by $\mathcal{V}(\cdot)$, producing frame-level representations $\mathbf{v}_t \in \mathbb{R}^{(1+N_p) \times C}$. Here, $T$ is the number of frames, $H$ and $W$ denote height and width, and $N_p$ is the number of visual patches.

For the skeleton modality, we extract skeleton data from each frame (\textbf{\textit{see Appendix A for details}}) to form $\mathbf{S} = \{ \mathbf{S}_t \}_{t=1}^{T}$, where $\mathbf{S}_t \in \mathbb{R}^{J \times 3}$ consists of $J$ joints and 3D coordinates $(x,y,z)$. The skeleton encoder $\mathcal{S}(\cdot)$ processes each frame to produce $\mathbf{s}_t \in \mathbb{R}^{(1+J) \times C}$(\textbf{\textit{see Appendix B}}).

\subsection{Stage 1: Contrasitive Skeleton-Image Pretraining} 

Stage 1 aims to align skeleton features with those from the frozen CLIP visual encoder through contrastive pretraining, producing well-aligned visual features rich in appearance cues and skeleton features that capture motion.

Specifically, paired skeleton–image sequences are processed by $\mathcal{V}(\cdot)$ and $\mathcal{S}(\cdot)$ to extract frame-level features, which are average-pooled across tokens and frames to obtain sequence-level representations: $\bar{\mathbf{v}} \in \mathbb{R}^C$ for the visual modality and $\bar{\mathbf{s}} \in \mathbb{R}^C$ for the skeleton modality. These representations are then used for contrastive pretraining.

By applying this operation to each sample, we obtain a multimodal representation set $\mathcal{M} = \left\{ (\bar{\mathbf{v}}_i, \bar{\mathbf{s}}_i) \right\}_{i=1}^{N_1}$, where $N_{1}$ denotes the number of sequence pairs in stage 1. The similarity between the two modalities is then computed as:

\begin{equation}
\text{Sim}\left(\bar{\mathbf{v}}_i, \bar{\mathbf{s}}_i\right) = \mathcal{J}_{v}\left(\bar{\mathbf{v}}_i\right) \cdot \mathcal{J}_{s}\left(\bar{\mathbf{s}}_i\right),
\label{similarity}
\end{equation}

where $\mathcal{J}_{v}$ and $\mathcal{J}_{s}$ are linear projections into a shared feature space. 
Similar to CLIP-ReID ~\cite{li2023clip}, \textbf{\textit{we adopt $\mathcal{L}_{v 2 s}$ and $\mathcal{L}_{s 2 v}$ as shown in Eq.\eqref{eq:loss_v2s}, Eq.\eqref{eq:loss_s2v} to align cross-modal features}}:

\begin{equation}
\mathcal{L}_{\text{v2s}}(i) = \frac{-1}{|\mathcal{P}_i|} \sum_{j \in \mathcal{P}_i} 
\log \frac{ \exp\left( \text{Sim}(\bar{\mathbf{v}}_i, \bar{\mathbf{s}}_j) / \tau \right) }
{ \sum\limits_{k=1}^{B} \exp\left( \text{Sim}(\bar{\mathbf{v}}_i, \bar{\mathbf{s}}_k) / \tau \right) },
\label{eq:loss_v2s}
\end{equation}

\begin{equation}
\mathcal{L}_{\text{s2v}}(i) = \frac{-1}{|\mathcal{P}_i|} \sum_{j \in \mathcal{P}_i} 
\log \frac{ \exp\left( \text{Sim}(\bar{\mathbf{v}}_j, \bar{\mathbf{s}}_i) / \tau \right) }
{ \sum\limits_{k=1}^{B} \exp\left( \text{Sim}(\bar{\mathbf{v}}_k, \bar{\mathbf{s}}_i) / \tau \right) }.
\label{eq:loss_s2v}
\end{equation}

Here, $\mathcal{P}_i = \{ j \mid y_i = y_j \}$ is the set of positive pairs that share the same identity label as the $i$-th sample. $\tau$ is a temperature hyperparameter that controls the sharpness of the distribution. 
$\mathcal{L}_{\text{v2s}}(i)$ and $\mathcal{L}_{\text{s2v}}(i)$ represent the supervised contrastive loss for aligning visual-to-skeleton and skeleton-to-visual representations, respectively.

\subsection{Stage 2: Prototype-guided Finetuning} 
Stage 2 employs prototype-guided finetuning to optimize the visual encoder, as ReID fundamentally relies on its ability to generate discriminative features for identity matching. 

\subsubsection{Prototype Fusion Updater.}

As shown in Fig.~\ref{fig:overview} (IV), the Prototype Fusion Updater (PFU) first combines visual and skeleton modalities to construct fusion prototypes, and then updates them using features within each training batch.

\textbf{Prototype Fusion.} First, we integrate aligned visual and skeleton features to produce more robust and discriminative fusion prototypes.
We \textbf{\textit{load the two encoders pretrained during stage 1 and freeze their parameters}} to ensure consistent feature extraction. 
Given aligned features $\{ \bar{\mathbf{v}}_i \}_{i=1}^{N_{1}}$ and $\{ \bar{\mathbf{s}}_i \}_{i=1}^{N_{1}}$ with identity labels $\{ y_i \}_{i=1}^{N_{1}}$, 
we compute modality-specific prototypes by averaging sequence-level features of all samples sharing the same identity, as illustrated by the Intra-ID pooling step in Fig.~\ref{fig:overview}; this step is performed only once during Stage 2.

\begin{equation}
P_{\text{S}}^{(c)} = \frac{1}{|\mathcal{I}_c|} \sum_{i \in \mathcal{I}_c} \bar{\mathbf{s}}_i, \quad
P_{\text{V}}^{(c)} = \frac{1}{|\mathcal{I}_c|} \sum_{i \in \mathcal{I}_c} \bar{\mathbf{v}}_i,
\end{equation}

where $\mathcal{I}_c = \{ i \mid y_i = c \}$ denotes the set of training samples of identity $c$. The modality-specific prototypes $P_{S}$ and $P_{V} \in \mathbb{R}^{K \times C}$ are then fused by an adaptive fusion mechanism, which learns a dynamic weight $\alpha \in \mathbb{R}^{K \times 1}$:

\begin{equation}
\alpha = \sigma \left( MLP \left( [P_{S} \, | \, P_{V}] \right) \right),
\label{eq:proto_weight}
\end{equation}

\begin{equation}
P_{F} = \alpha P_{S} + (1 - \alpha) P_{V},
\label{eq:proto_fusion}
\end{equation}

Here, $\alpha \in \mathbb{R}^{K \times 1}$ is an adaptive weight for each class, learned from concatenated modality features. The symbol $[\, \cdot \mid \cdot \,]$ denotes feature-wise concatenation, MLP is a two-layer fully connected network, followed by a sigmoid activation $\sigma(\cdot)$ to constrain the output to $(0,1)$.
This design enables \textbf{\textit{class-aware fusion by dynamically adjusting each modality’s contribution}}, allowing the final prototypes to comprehensively capture both discriminative appearance cues and inherent structural patterns.

\textbf{Prototype Update.} Secondly, we observe that using fixed fusion prototypes limits adaptability as it \textbf{\textit{overlooks the appearance diversity within the same identity}}. Therefore, we dynamically adjust the fusion prototypes for each input sequence to capture sequence-specific characteristics.

For each training batch, we extract visual features 
$ \mathbf{f}_{v} $ from the visual encoder 
and skeleton features $ \mathbf{f}_{\text{s}}$ from the skeleton encoder.
They are concatenated along the token dimension to form a fused sequence $ F \in \mathbb{R}^{B \times (L_{\text{vis}} + L_{\text{ske}}) \times C} $, where $B$ is the batch size, $L_{\text{vis}}=1+N_{p}$ and $L_{\text{ske}}=1+J$ are the numbers of visual and skeleton tokens.

This multimodal sequence encodes both appearance and structural cues and serves as the key and value for a cross-attention module.
Meanwhile, batch-wise fusion prototypes $ P_{F} \in \mathbb{R}^{B \times K \times C} $, representing $K$ identity prototypes, act as queries.
PFU adopts a transformer-style architecture comprising self-attention, cross-attention, and a feed-forward MLP. The update process is formulated as:

\begin{equation}
\hat{P}_F = P_F + \mathrm{MLP}\big( \mathrm{CrossAttn} ( \mathrm{SelfAttn}(P_F), F ) \big).
\end{equation}

The update begins with $\mathrm{SelfAttn}(P)$ over prototype tokens to enable inter-prototype interaction. The result then attends to the fused tokens $F$ through cross-attention to \textbf{\textit{capture sample-specific details from multi-modal context}}. Finally, a feed-forward network $\mathrm{MLP}(\cdot)$ refines the output, which is added back through a residual connection to produce the updated prototypes $\hat{P}_F$.

\textbf{Prototype Supervision Loss.} Then the updated fusion prototypes $\hat{P}_F$ are used to supervise ReID. Given prototypes $\hat{P}_1, \dots, \hat{P}_K$ and a visual feature $f_i$ from the $i$-th training sample, the classification loss $\mathcal{L}_{CSIP}$ is defined as:

\begin{equation}
\mathcal{L}_{CSIP}(i) = - \sum_{k=1}^{K} q_k \log 
\frac{\exp \left( f_i^\top \hat{P}_k \right)}
{\sum_{j=1}^{K} \exp \left( f_i^\top \hat{P}_j \right)}.
\end{equation}

What’s more, we follow the strong pipeline~\cite{luo2019bag} and adopt both the cross-entropy loss $\mathcal{L}_{\text{CE}}$ with label smoothing and the triplet loss $\mathcal{L}_{\text{Triplet}}$ to jointly optimize the visual encoder.

\subsubsection{Skeleton Guided Temporal Modeling.}

To model temporal dynamics, we propose Skeleton-Guided Temporal Modeling (SGTM) module as shown in Fig.~\ref{fig:overview} (V).

\textbf{Message Token Encoding.}
Given $X^{\text{vis}} \in \mathbb{R}^{T \times (1+N_p) \times C}$, we average all tokens per $t$ to extract a compact message token.
Unlike directly using [CLS] token, we adopt average pooling since recent work ~\cite{he2022transfg} has shown that \textbf{\textit{patch tokens still retains rich semantic information}}. 
Pooled tokens are projected via $W_v$ into a shared space and enhanced by temporal MHSA:

\begin{equation}
\tilde{\mathbf{m}}^{\text{vis}} = 
\text{MHSA}\big( W_v \,\mathrm{Pool}(X^{\text{vis}}_t) \big),
\end{equation}

where $W_v$ is a learnable linear projection, $\mathrm{Pool}(\cdot)$ denotes average pooling over all tokens at $t$, and $\mathrm{MHSA}(\cdot)$ refers to temporal self-attention. Skeleton message tokens $\tilde{\mathbf{m}}^{\text{ske}}$ are computed in the same manner.

\textbf{Auxiliary Temporal Distillation.}
Following the Learning Using Privileged Information (LUPI) paradigm~\cite{vapnik2009new}, we \textbf{\textit{leverage skeleton features as privileged information available only during training}}. ATD employs cross-attention to \textbf{\textit{distill skeleton-guided motion cues}} into visual message tokens:

\begin{equation}
\hat{\mathbf{m}}^{\text{vis}} = \text{CrossAttn}(\tilde{\mathbf{m}}^{\text{vis}}, \tilde{\mathbf{m}}^{\text{ske}}, \tilde{\mathbf{m}}^{\text{ske}}),
\end{equation}

To enhance modality awareness, we inject learnable type embeddings $\mathbf{E} \in \mathbb{R}^{4 \times C}$ for four token types: $\{\mathbf{x}_{t,i}^{\text{vis}}, \mathbf{\hat{m}}^{\text{vis}}, \mathbf{x}_{t,j}^{\text{ske}}, \mathbf{\tilde{m}}^{\text{ske}}\}$, enabling explicit source differentiation.
Such a design has proven effective in~\cite{devlin2019bert}.
By distilling temporal cues from skeleton into visual features under the LUPI framework, ATD strengthens the temporal modeling capacity of the visual stream while keeping inference free from skeleton data.

\textbf{Temporal Aggregation.}
TA integrates temporal dependencies across token types by forming a unified sequence $\mathbf{X} \in \mathbb{R}^{L \times BT \times C}$, where $L$ varies with training (which includes all four token types) 
and messages $\tilde{\mathbf{m}}^{\text{ske}}$) and testing (which includes only visual tokens $\mathbf{x}_{t,i}^{\text{vis}}$ and $\hat{\mathbf{m}}^{\text{vis}}$):

\begin{equation}
\mathbf{X} =
\begin{cases}
\left[ \mathbf{x}_{t,i}^{\text{vis}} \,\|\, \hat{\mathbf{m}}^{\text{vis}} \,\|\, \mathbf{x}_{t,j}^{\text{ske}} \,\|\, \tilde{\mathbf{m}}^{\text{ske}} \right], & \text{if training} \\
\left[ \mathbf{x}_{t,i}^{\text{vis}} \,\|\, \hat{\mathbf{m}}^{\text{vis}} \right], & \text{if testing}
\end{cases}
\label{eq:TA}
\end{equation}

The unified sequence $\mathbf{X}$ is fed into an attention block as shown in Fig.~\ref{fig:overview} (V), comprising a temporal self-attention layer followed by a feed-forward network, both equipped with residual connections and layer normalization.

\textbf{Frame-level Supervision Loss.}
We apply attention-based pooling over tokens to obtain frame-level logits $\mathbf{z}_{i,t} \in \mathbb{R}^C$, where the attention weights highlight informative tokens and aggregate temporal context into a global representation. The classification loss is then computed as:

\begin{equation}
\mathcal{L}_{\text{Frame}} = - \sum_{i=1}^{B} \sum_{t=1}^{T} \sum_{k=1}^{K} q_{i,t,k} \log p_{i,t,k},
\end{equation}

where $K$ is the number of identity classes, $p_{i,t,k}$ is the softmax probability derived from $\mathbf{z}_{i,t}$, and $q_{i,t,k}$ is the corresponding frame-level label. This loss \textbf{\textit{enforces consistent identity predictions across all frames of a sample}}, enhancing frame-level discriminability and compensating for the reliance on sequence-level features in Stage 1 and PFU.

\subsection{Traing Strategy}

Our training strategy consists of two stages: Contrastive Skeleton-Image Pretraining and Prototype-guided Finetuning. 
In Stage 1, we load pretrained weights from CLIP, freeze the visual encoder and \textbf{\textit{optimize only the skeleton encoder}} to align the two modalities via supervised contrastive learning. The training objective is:
\begin{equation}
\mathcal{L}_{\text{stage } 1} = \mathcal{L}_{\text{v2s}} + \mathcal{L}_{\text{s2v}}.
\end{equation}

In Stage 2, we \textbf{\textit{jointly optimize the visual encoder, Prototype Fusion Updater (PFU), and Skeleton-Guided Temporal Modeling (SGTM)}}. Specifically, we employ the cross-entropy loss $\mathcal{L}_{\text{CE}}$, triplet loss $\mathcal{L}_{\text{Triplet}}$, prototype-guided supervision loss $\mathcal{L}_{\text{CSIP}}$ from PFU, and frame-level supervision loss $\mathcal{L}_{\text{Frame}}$ from SGTM. Two hyperparameters $\lambda_1$ and $\lambda_2$ control the contribution of the last two terms:

\begin{equation}
\mathcal{L}_{\text{stage } 2} = \mathcal{L}_{\text{CE}} + \mathcal{L}_{\text{Triplet}} + \lambda_1 \mathcal{L}_{\text{CSIP}} + \lambda_2 \mathcal{L}_{\text{Frame}}.
\end{equation}

The whole training process of the proposed CSIP-ReID is summarized in \textbf{\textit{Algorithm 1 (see Appendix C)}}.

\section{Experiments}

\subsection{Datasets and Evaluation Protocols}
We evaluate CSIP-ReID on three video-based person ReID benchmarks: MARS~\cite{zheng2016mars}, LS-VID~\cite{li2019global}, and iLIDS-VID~\cite{wang2014person}, \textbf{\textit{details provided in Appendix D}}.
Following common practice, we evaluate model performance using the Cumulative Matching Characteristic (CMC) curve at Rank-k and mean Average Precision (mAP)~\cite{zheng2015scalable}.

\subsection{Experiment Settings}
Our model is implemented in PyTorch and trained on a single NVIDIA Tesla L20 GPU.
We sample 8 frames per tracklet, resize them to 256$\times$128, and apply data augmentation as in TF-CLIP~\cite{tfclip_2024}. Stage 1 is trained for 120 epochs with a batch size of 64, while Stage 2 is trained for 80 epochs using PK sampling~\cite{hermans2017defense} with 4 identities $\times$ 4 tracklets. 
The skeleton encoder parameters follow those in TranSG~\cite{rao2023transg}. Detailed settings are provided in \textbf{\textit{Appendix D}}. Code is available in https://github.com/Rifen-Lin/CSIP-ReID.git.

% resizebox
\begin{table*}[]
    \centering
    \resizebox{0.90\textwidth}{!}{%
        \begin{tabular}{cccccccc}
\hline
\multirow{2}{*}{Methods} & \multirow{2}{*}{Source} & \multicolumn{2}{c}{MARS} & \multicolumn{2}{c}{LS-VID} & \multicolumn{2}{c}{iLIDS-VID} \\
                         &                         & mAP & Rank-1               & mAP & Rank-1                & Rank-1 & Rank-5                \\ \hline
STMP~\cite{liu2019spatial}         & AAAI19   & 72.7  & 84.4  & 39.1  & 56.8  & 84.3  & 96.8  \\
M3D~\cite{li2019multi}             & AAAI19   & 74.1  & 84.4  & 40.1  & 57.7  & 74.0  & 94.3  \\
GLTR~\cite{li2019global}           & ICCV19   & 78.5  & 87.0  & 44.3  & 63.1  & 86.0  & 98.0  \\
TCLNet~\cite{hou2020temporal}      & ECCV20   & 85.1  & 89.8  & 70.3  & 81.5  & 86.6  & -     \\
MGH~\cite{yan2020learning}         & CVPR20   & 85.8  & 90.0  & 61.8  & 79.6  & 85.6  & 97.1  \\
BiCnet-TKS~\cite{hou2021bicnet}    & CVPR21   & 86.0  & 90.2  & 75.1  & 84.6  & -     & -     \\
CTL~\cite{liu2021spatial}          & CVPR21   & 86.7  & 91.4  & -     & -     & 89.7  & 97.0  \\
DIL~\cite{he2021dense}             & ICCV21   & 87.0  & 90.8  & -     & -     & 92.0  & 98.0  \\
CAVIT~\cite{wu2022cavit}           & ECCV22   & 87.2  & 90.8  & 79.2  & 89.2  & 93.3  & 98.0  \\
SINet~\cite{bai2022salient}        & CVPR22   & 86.2  & 91.0  & 79.6  & 87.4  & 92.5  & -     \\
SDCL~\cite{cao2023event}          & CVPR23  & 86.5  & 91.1  & -     & -     & -  & 93.2  \\
TCVIT~\cite{wu2024temporal} & AAAI24 & 87.6 & 91.7 & 83.1 & 90.1 & 94.3 & \underline{99.3}  \\
TF-CLIP~\cite{tfclip_2024} & AAAI24 & 89.4 & 93.0 & 83.8 & 90.4 & 94.5  & 99.1  \\
CLIMB-ReID~\cite{Yu_Liu_Zhu_Wang_Zhang_Lu_2025} & AAAI25 & 89.7 & \underline{93.3} & \textbf{85.0} & \underline{91.3} & \underline{96.7} & \textbf{99.9} \\
\textbf{CSIP-ReID(Ours)} &                    & \textbf{90.4} & \textbf{94.2} & \underline{84.2} & \textbf{92.5} & \textbf{97.2} & 98.2 \\ 
\hline
\end{tabular}
        }
    \caption{Comparison with typical methods on MARS, LS-VID and iLIDS-VID.
    Please refer to Appendix D for the full table.
    }
    \label{tab:Sota1}
\end{table*}

\begin{table*}[t]
    \begin{center}
        \resizebox{0.92\textwidth}{!}
        {
                \begin{tabular}{cccc|cc|ccc|ccc}
	\hline
	\multirow{2}{*}{Model} 
        & \multirow{2}{*}{\begin{tabular}{c}Prototype\\Fusion\end{tabular}}
        & \multirow{2}{*}{\begin{tabular}{c}Prototype\\Update\end{tabular}}
        & \multirow{2}{*}{SGTM} 
	& \multirow{2}{*}{Params(M)} 
	& \multirow{2}{*}{FLOPs(G)} 
	& \multicolumn{3}{c|}{MARS} 
	& \multicolumn{3}{c}{LS-VID} \\
	& & & 
	& & & mAP & Rank-1 & Rank-5 & mAP & Rank-1 & Rank-5 \\ \hline
        1 &  $\times$ & $\times$ & $\times$ & 86.95 & 16.98 & 85.6 & 90.4 & 96.4 & 80.2 & 87.2 & 95.5 \\
	2 &  $\times$ & $\times$ & $\times$ & 107.15 & 16.99 & 88.3 & 90.6 & 96.9 & 80.0 & 87.9 & 95.6 \\
	3 &  \checkmark & $\times$ & $\times$ & 109.78 & 20.34 & 88.4 & 92.3 & 97.8 & 82.8 & 91.0 & 97.1 \\
	4 &  \checkmark & \checkmark & $\times$ & 118.98 & 20.49 & 89.2 & 92.5 & \underline{98.0} & 82.9 & 91.1 & 97.1 \\
	5 &  \checkmark & $\times$ & \checkmark & 125.91 & 21.13 & \underline{90.1} & \underline{93.4} & \underline{98.0} & \underline{83.4} & \underline{91.5} & \underline{97.2} \\
	6 &  \checkmark & \checkmark & \checkmark & 135.11 & 21.28 & \textbf{90.4} & \textbf{94.2} & \textbf{98.3} & \textbf{84.2} & \textbf{92.5} & \textbf{97.3} \\ \hline
\end{tabular}
        }
    \end{center}
    \caption{Comparison of different modules and the computational cost on MARS and LS-VID. 
    }
    \label{table:ablation}
\end{table*}

\subsection{Comparison with State-of-the-arts}

We compare our method with state-of-the-art approaches on three video-based person ReID benchmarks, with results shown in Tab.~\ref{tab:Sota1}, demonstrating its superior performance.

\textbf{\textit{On MARS}}, CSIP-ReID achieves the best performance with an mAP of 90.4\% and Rank-1 accuracy of 94.2\%. It surpasses TF-CLIP by 1.0\% in mAP and 1.2\% in Rank-1, largely because CSIP-ReID leverages skeleton guidance as a complementary second modality.
\textbf{\textit{On LS-VID}}, CSIP-ReID achieves the best Rank-1 accuracy on LS-VID at 92.5\%, surpassing TF-CLIP by 2.1\%, and ranks second in mAP, slightly behind CLIMB-ReID, likely due to a broader similarity neighborhood that admits a few hard negatives, which can be mitigated by re-ranking.
\textbf{\textit{On iLIDS-VID}}, CSIP-ReID attains the best Rank-1 accuracy on iLIDS-VID at 97.2\%, exceeding CLIMB-ReID by 0.5\% and TF-CLIP by 2.7\%. 
The Rank-5 accuracy is 98.2\%, slightly below CLIMB-ReID, likely because the small scale of this dataset limits contrastive pretraining in Stage1.

\subsection{Ablation Study}
To evaluate the contribution of each component in our model, we conduct ablation studies on the MARS and LS-VID datasets, with results summarized in Table~\ref{table:ablation}.
$Model1$ serves as the baseline, where only the CLIP vision encoder is fine-tuned.
$Model2$ refers to TF-CLIP without the Sequence-Specific Prompt and Temporal Memory Diffusion modules, where the visual encoder is fine-tuned solely under the guidance of the visual prototype.

\subsubsection{Effectiveness of Prototype Fusion.}
As shown in the first three rows of Table~\ref{table:ablation}, $Model2$ outperforms $Model1$ on MARS by leveraging visual prototypes, while $Model3$ achieves further gains by jointly using visual and skeleton prototypes, \textbf{\textit{confirming their complementarity}}. 
Similar trends are observed on LS-VID, demonstrating that fusion prototypes provide more effective guidance for fine-tuning the visual encoder than visual prototypes alone.

\subsubsection{Effectiveness of Prototype Update.}
As shown in Tab.~\ref{table:ablation}, compared with $Model3$, adding prototype updates of PFU to $Model4$ improves mAP by 0.8\% and Rank-1 accuracy by 0.2\% on MARS, with similar gains on LS-VID, demonstrating its effectiveness.
This improvement likely stems from online prototype updates of PFU, which adapt to each batch and \textbf{\textit{capture fine-grained details overlooked by static fusion prototypes}}, yielding more discriminative representations.

\subsubsection{Effectiveness of SGTM.}
As shown in Tab.~\ref{table:ablation}, the proposed SGTM significantly improves performance, with $Model5$ achieving gains of 1.7\% mAP and 1.1\% Rank-1 accuracy over $Model3$ on MARS, and similar improvements on LS-VID.
These results highlight SGTM’s effectiveness, as it distills additional temporal cues from skeleton to enhance visual temporal modeling.

\begin{figure*}
	\centering
	\includegraphics[width=1.0\linewidth]{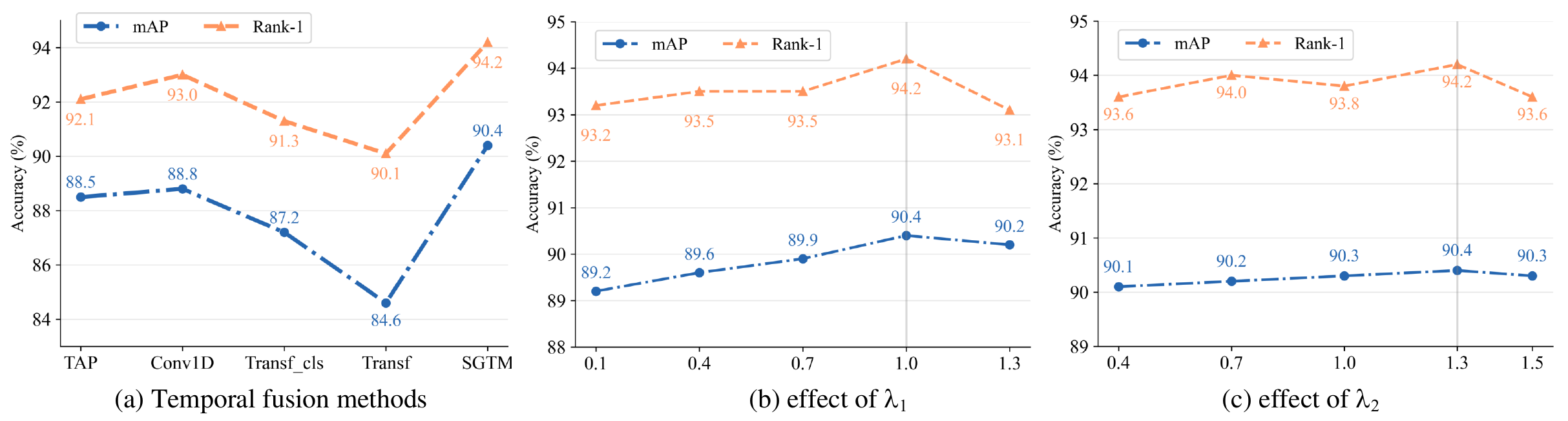}
	\caption{Analysis of key modules/factors affecting performance. This figure illustrates (a) the impact of different temporal fusion methods, (b) the effect of the hyperparameter $\lambda_1$, and (c) the effect of $\lambda_2$ on model performance.
	}
	\label{fig:balation}
\end{figure*}

\begin{figure}
	\centering
        \includegraphics[width=0.95\linewidth]{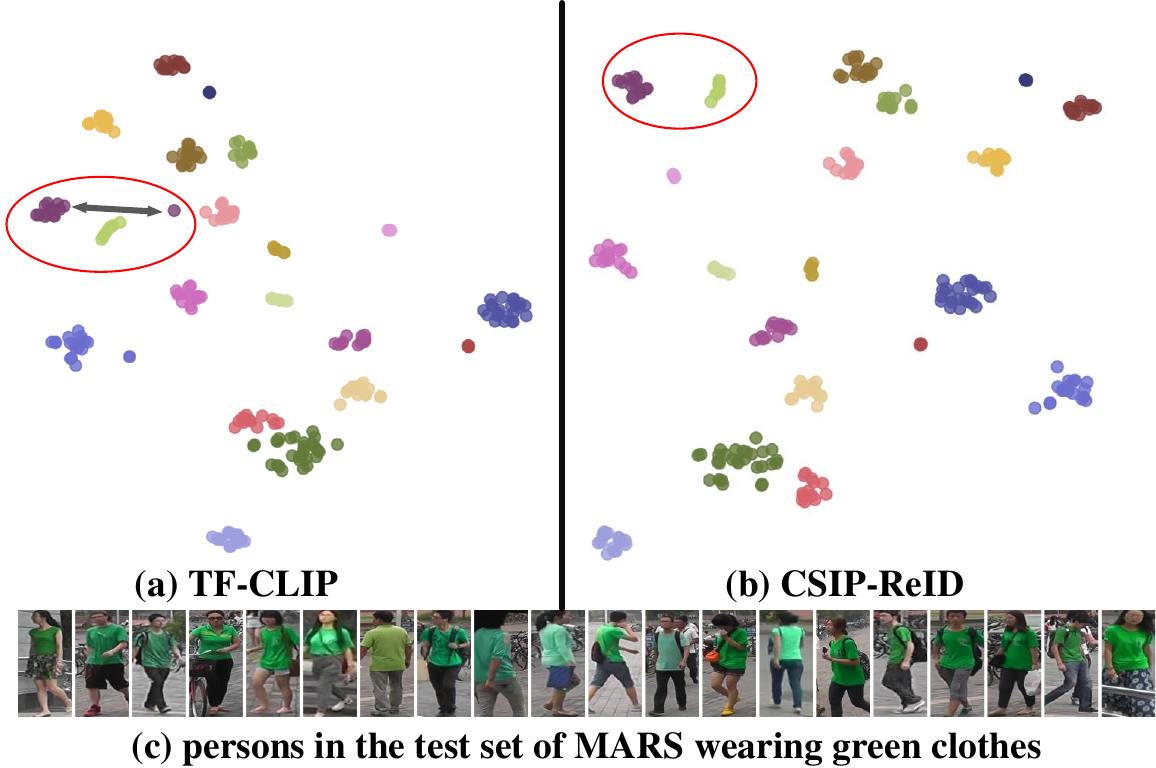}
	\caption{CSIP-ReID produces more compact, discriminative clusters than TF-CLIP in the t-SNE visualization. Each color represents a different identity. Red circles highlight samples from two visually similar identities.
    }
	\label{fig:vis_tSNE}
\end{figure}

\subsubsection{Comparison of different temporal fusion methods.}
To assess different temporal aggregation strategies, we compare several fusion methods on MARS, following TF-CLIP~\cite{tfclip_2024}.
As shown in Fig.~\ref{fig:balation}(a), SGTM achieves 94.2\% Rank-1 accuracy, surpassing the second-best method, Conv1D, by 1.2\%.
This improvement stems from SGTM’s ability to model temporal dynamics in visual frames while \textbf{\textit{distilling complementary cues from skeletons}}.

\begin{figure}
	\centering
	\includegraphics[width=0.85\linewidth]{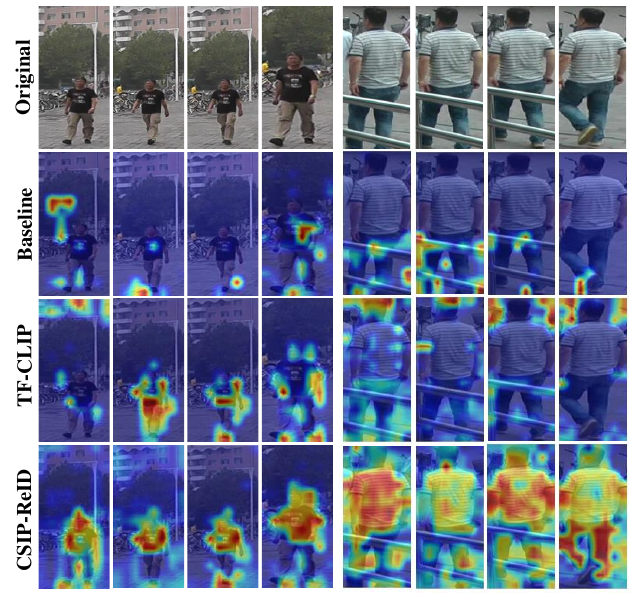}
	\caption{CSIP-ReID shows stronger attention focus on tokens corresponding to human regions. We compare the proposed method CSIP-ReID with Baseline and TF-CLIP. 
	}
	\label{fig:vis_attn}
\end{figure}

\subsubsection{The effect of $\lambda_1$.}
In stage 2, the parameter $\lambda_1$ controls the weight of the prototype supervision loss $\mathcal{L}_{\text{CSIP}}$, which \textbf{\textit{enhances identity discrimination through joint visual–skeleton representation}}.
As shown in Fig.~\ref{fig:balation}(b), the model achieves optimal Rank-1 accuracy and mAP at $\lambda_1 = 1.0$. A smaller $\lambda_1$ weakens prototype supervision, while a larger value causes overfitting to prototypes and distorts the video feature space. Thus, \textbf{$\lambda_1 = 1.0$} is adopted for balanced cross-modal supervision and optimal performance.

\subsubsection{The effect of $\lambda_2$.}
The parameter $\lambda_2$ controls the weight of frame-level supervision loss $\mathcal{L}_{\text{Frame}}$, which \textbf{\textit{enforces fine-grained alignment of individual frame features}}.
As shown in Fig.~\ref{fig:balation}(c), performance improves as $\lambda_2$ increases, peaking at 1.3, where temporal discrimination is best. Higher values may cause overfitting to frame-specific noise. Thus, we set $\lambda_2 = 1.3$ to balance precision and generalization.

\subsection{Visualization}

\subsubsection{t-SNE visualization.}
To demonstrate the effect of CSIP-ReID, we visualize t-SNE~\cite{van2008visualizing} distributions of visually similar pedestrians.
In Fig.~\ref{fig:vis_tSNE}(a) and (b), each color represents an identity, with red circles marking two similar ones.
Compared with TF-CLIP which exhibits scattered features and outliers, CSIP-ReID \textbf{\textit{forms more compact and separable clusters}}, reducing intra-person variance and enhancing inter-person separation.

\subsubsection{Focus Region Analysis.}
Finally, we visualize CAM results in Fig.~\ref{fig:vis_attn}, where warmer colors indicate stronger identity-related attention. 
The baseline focuses on local cues and fails to capture holistic identity information. 
TF-CLIP improves attention with visual prototypes and temporal memory but still occasionally attends to background tokens. In contrast, our method \textbf{\textit{leverages skeleton guidance to direct attention primarily toward human regions}}, yielding more identity-relevant focus.

\subsection{Transfer to Skeleton-based ReID}

To further evaluate the generalization ability of CSIP-ReID, we extend it to skeleton-based ReID as shown in Tab.~\ref{table:skeleton_based_results}. 
The model architecture, experimental settings and additional results on BIWI~\cite{munaro2014one} and IAS~\cite{munaro2014feature}, are \textbf{\textit{detailed in Appendix D}}. 
Importantly, visual information is used only during training for cross-modal guidance and excluded during testing to ensure fair comparison, where CSIP-ReID consistently outperforms state-of-the-art approaches, \textbf{\textit{demonstrating strong generalization ability}}.

\begin{table}[t]
    \centering
    \resizebox{\linewidth}{!}{

\begin{tabular}{l|cc|cc|cc|cc}
\hline
\multirow{2}{*}{Methods} & \multicolumn{2}{c|}{BIWI-S} & \multicolumn{2}{c|}{BIWI-W} & \multicolumn{2}{c|}{IAS-A} & \multicolumn{2}{c}{IAS-B} \\
 & mAP & Rank-1 & mAP & Rank-1 & mAP & Rank-1 & mAP & Rank-1 \\
\hline
PoseGait~\shortcite{liao2020model} & 9.9 & 14.0 & 11.1 & 8.8 & 17.5 & 28.4 & 20.8 & 28.9 \\
SGELA~\shortcite{rao2021self} & 15.1 & 25.8 & 19.0 & 11.7 & 13.2 & 16.7 & 14.0 & 22.2 \\
SimMC~\shortcite{rao2022simmc} & 12.3 & 41.7 & 19.9 & 24.5 & 18.7 & 44.8 & 22.9 & 46.3 \\
Hi-MPC \shortcite{rao2024hierarchical} & 17.4 & 47.5 & 22.6 & 27.3 & 23.2 & 45.6 & 25.3 & 48.2 \\
TranSG~\shortcite{rao2023transg} & 30.1 & \underline{68.7} & 26.9 & 32.7 & 32.8 & 49.2 & 39.4 & 59.1 \\
MoCos~\shortcite{rao2025motif} & \underline{32.1} & \textbf{72.0} & \underline{30.5} & \underline{36.0} & \underline{35.8} & \underline{51.9} & \underline{45.5} & \underline{61.5} \\
\textbf{CSIP-ReID} & \textbf{34.5} & 68.6 & \textbf{33.8} & \textbf{36.9} & \textbf{48.1} & \textbf{53.6} & \textbf{50.7} & \textbf{63.3} \\
\hline
\end{tabular}
            }
    \caption{skeleton-based performance comparison with typical methods. Please refer to Appendix D for the full table.}
    \label{table:skeleton_based_results}
\end{table}
\section{Conclusion}

In this paper, we explore the potential of skeleton-image pretraining to enhance video-based person ReID.
Specifically, we propose a novel two-stage framework named CSIP-ReID. 
Stage 1 aligns visual and skeleton features using supervised contrastive loss, while Stage 2 introduces a Prototype Fusion Updater (PFU) to fuse motion and appearance cues.
% To further capture temporal dynamics, we design a Skeleton Guided Temporal Modeling (SGTM) module, which distills temporal cues from skeleton modality. 
A Skeleton-Guided Temporal Modeling (SGTM) module further distills temporal information from the skeleton modality.
Experiments on three benchmarks demonstrate the effectiveness of CSIP-ReID, and its transfer to skeleton-based ReID highlights strong generalization.

% Despite its advantages, CSIP-ReID incurs additional costs for skeleton extraction when the dataset lacks this modality. Meanwhile, the pretraining is performed separately on each dataset, resulting in pretrained weights tailored only to dataset-specific feature distributions.
% Future work will explore pretraining across multiple ReID datasets to further improve skeleton-image sequence alignment.

\section{Acknowledgments}
This work was supported in part by the High Performance Computing Center of Central South University.

\bibliography{aaai2026}

\newpage
\section{Skeleton Data Extraction}

We develop a two-step pipeline to extract 3D skeleton data from video frames in our ReID dataset. The process employs the \textbf{\textit{HMR2.0 (Human Mesh Recovery 2.0)}} model~\cite{goel2023humans} to reconstruct detailed 3D human meshes and subsequently applies a regression step to obtain compact skeletal representations.

In the first step, the pre-trained HMR2.0 model is applied to each frame to recover a triangulated SMPL mesh. Given image sequences grouped by person ID, HMR2.0 predicts high-resolution meshes consisting of 6890 vertices per frame, which are stored as \textbf{\textit{.obj}} files.

In the second step, these SMPL meshes are converted into 3D skeletons. The conversion script parses each \textbf{\textit{.obj}} file to extract vertex coordinates and applies a fixed joint regressor 
\( J_{reg} \in \mathbb{R}^{17 \times 6890} \) 
to compute 17 canonical joints. This regressor performs a linear mapping according to  
\[
J = J_{reg} \cdot V,
\]
where \( V \in \mathbb{R}^{6890 \times 3} \) is the vertex matrix and \( J \in \mathbb{R}^{17 \times 3} \) represents the resulting joint coordinates. The extracted joints follow the Human3.6M convention, preserving essential body structure while significantly reducing data dimensionality. Each frame’s joints are saved as JSON files with 17 three-dimensional keypoints, while intermediate \textbf{\textit{.obj}} files are automatically removed to minimize storage.

To enable large-scale processing, the entire workflow is automated with a Bash script that iterates over all person IDs, executes HMR2.0 inference, and performs mesh-to-skeleton conversion. The automation checks for existing outputs and skips completed samples, ensuring efficient and reproducible extraction. As a result, raw video frames are transformed into lightweight, standardized 3D skeleton data suitable for downstream person ReID modeling.

\begin{figure*}[t]
	\centering
	\includegraphics[width=1.0\textwidth]{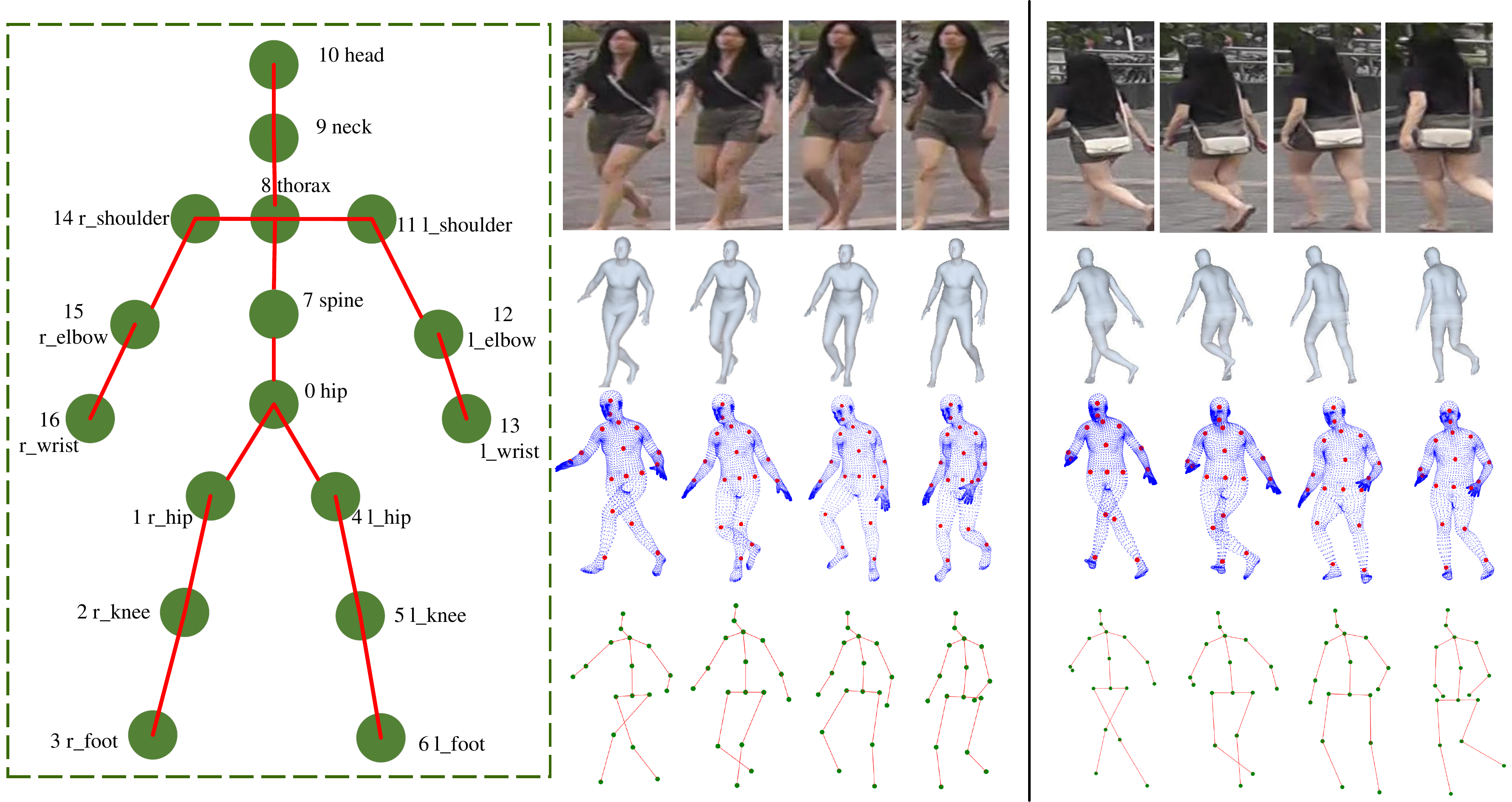}
	\caption{\textbf{Overview of the skeleton data extraction pipeline}. The left panel illustrates the Human3.6M joint definition consisting of 17 anatomical landmarks. The right panels show the sequential transformation from raw video frames to 3D skeletons: HMR2.0 first reconstructs high-resolution SMPL meshes from RGB frames, 
    after which a joint regressor maps the mesh vertices to 17 canonical 3D joints. The bottom row demonstrates the resulting lightweight skeleton graphs, which preserve the essential body structure while significantly reducing data dimensionality.}

	\label{fig:skeleton_extraction}
\end{figure*}

Figure~\ref{fig:skeleton_extraction} provides a visual overview of this pipeline. The left panel shows the Human3.6M joint definition with 17 anatomical landmarks, serving as the target skeleton representation. The right panel depicts the stepwise transformation from raw RGB frames to 3D skeletons: the first row shows the original input frames; the second row presents the SMPL meshes reconstructed by HMR2.0; the third row overlays the regressed joints on the meshes; and the final row visualizes the extracted joints as lightweight skeleton graphs. This example highlights how our pipeline converts unstructured RGB sequences into standardized 3D skeletons for effective ReID modeling.

\section{Details of Skeleton Graph transformer}

For the skeleton modality, the extracted skeleton data from each frame are organized as a sequence 
\(\mathbf{S} = \{ \mathbf{S}_t \}_{t=1}^{T} \), 
where each \(\mathbf{S}_t \in \mathbb{R}^{J \times 3}\) is modeled as a spatial graph with joints as nodes and bones as edges. 
The Skeleton Graph Transformer (SGT) \(\mathcal{S}(\cdot)\) takes this sequence as input and captures both structural and motion-related dependencies by learning full relations among all body joints. 
Unlike traditional GCNs that only aggregate information from local neighborhoods, SGT adopts a multi-head self-attention mechanism that allows every joint to attend to all others, and further incorporates Laplacian positional encoding to inject structural priors.  
Formally:
\begin{equation}
\mathbf{S}_t = \text{SGT}\left( \mathcal{G}^{(t)}; \mathbf{A}, \mathbf{E}_{\text{pos}}^{\text{ske}} \right) = \big[ \mathbf{h}_1^{(t)}, \mathbf{h}_2^{(t)}, \dots, \mathbf{h}_J^{(t)} \big],
\label{skeleton_encoder}
\end{equation}
where \(\mathbf{s}_t \in \mathbb{R}^{J \times C}\) denotes the joint features of frame \(t\), \(J\) is the number of joints, and \(C\) is the hidden feature dimension. 
The SGT processes the skeleton graph \(\mathcal{G}^{(t)}\) with adjacency matrix \(\mathbf{A}\) and positional encoding \(\mathbf{E}_{\text{pos}}^{\text{ske}}\), yielding frame-level representations \(\mathbf{S}_{\text{frame}} = \{ \mathbf{s}_t \}_{t=1}^{T}\).

The SGT workflow can be divided into four components: Graph Embedding, Full-Relation Attention, Graph Prototype Contrastive Loss (\( \mathcal{L}_{\text{GPC}} \)), Structure-Trajectory Prompted Reconstruction Loss (\( \mathcal{L}_{\text{STPR}} \)), and the Final Training Objective.

\paragraph{Graph Embedding.}
Given a skeleton sequence \(\mathbf{S} \in \mathbb{R}^{T \times J \times 3}\), we first embed the 3D coordinates of each joint into a \(C\)-dimensional space and add joint-specific positional encoding obtained via Laplacian eigenmaps:
\begin{equation}
\mathbf{h}_{t,j}^{(0)} = \text{FC}_2\Big( \text{ReLU}\big(\text{FC}_1(\mathbf{S}_{t,j})\big) \Big) + \text{FC}_{\text{pos}}(\mathbf{e}_j),
\end{equation}
where \(\mathbf{S}_{t,j} \in \mathbb{R}^3\) is the coordinate of the \(j\)-th joint at frame \(t\), and \(\mathbf{e}_j \in \mathbb{R}^k\) is its positional encoding.

\paragraph{Full-Relation Attention.}
For each frame \(t\), the node features \(\mathbf{H}^{(l-1)}_t \in \mathbb{R}^{J \times C}\) are updated by multi-head self-attention without adjacency masking:
\begin{equation}
\mathbf{H}^{(l)}_t = \mathcal{F}\big( \mathbf{H}^{(l-1)}_t + \text{MultiHeadAttn}(\mathbf{H}^{(l-1)}_t) \big),
\end{equation}
where
\begin{equation}
\text{Attention}^{(h)}(\mathbf{Q}, \mathbf{K}, \mathbf{V}) = \text{softmax}\Big( \frac{\mathbf{QK}^\top}{\sqrt{d}} \Big) \mathbf{V}.
\end{equation}
Here, \(h\) indexes the attention head, \(d\) is the scaling factor, and \(\mathcal{F}(\cdot)\) is a feed-forward network with residual connections. After \(L\) layers, joint-level features \(\mathbf{H}_t^{(L)} \in \mathbb{R}^{J \times C}\) are obtained. The frame-level feature is computed by averaging node features:
\begin{equation}
\mathbf{s}_t = \frac{1}{J} \sum_{j=1}^J \mathbf{h}_{t,j}^{(L)}.
\end{equation}
Finally, sequence-level representation is derived via temporal average pooling:
\begin{equation}
\mathbf{S}_{\text{seq}} = \frac{1}{T} \sum_{t=1}^T \mathbf{s}_t.
\end{equation}

\paragraph{Graph Prototype Contrastive Loss (\( \mathcal{L}_{\text{GPC}} \)).}
To enhance identity discrimination, SGT adopts a \textbf{Graph Prototype Contrastive Loss} that pulls skeleton features toward their class prototypes while pushing them away from other classes. 
For each identity \(k\), the prototype \(c_k\) is defined as the centroid of its sequence-level features. 
The overall loss is a weighted combination of sequence-level and skeleton-level terms:
\begin{equation}
\mathcal{L}_{\text{GPC}} = \alpha \mathcal{L}_{\text{seq}} + (1 - \alpha) \mathcal{L}_{\text{ske}}.
\end{equation}

The sequence-level term is formulated as:
\begin{equation}
\mathcal{L}_{\text{seq}} = \frac{1}{N} 
\sum_{(k,j) \in \mathcal{I}_s} 
-\log \frac{\exp( z_{k,j,k} / \tau_1 )}
{\sum_{m=1}^{C} \exp( z_{k,j,m} / \tau_1 )},
\end{equation}
where \( z_{k,j,m} = \mathbf{S}_{k,j} \cdot c_m \) measures the similarity between sequence feature \(\mathbf{S}_{k,j}\) and prototype \(c_m\), and \(\mathcal{I}_s\) indexes all training sequences.

Similarly, the skeleton-level term is:
\begin{equation}
\mathcal{L}_{\text{ske}} = \frac{1}{TN} 
\sum_{(k,j,t) \in \mathcal{I}} 
-\log \frac{\exp( z_{t,k,j,k} / \tau_2 )}
{\sum_{m=1}^{C} \exp( z_{t,k,j,m} / \tau_2 )},
\end{equation}
where \( z_{t,k,j,m} = F_1(\mathbf{s}_{t,k,j}) \cdot F_2(c_m) \) is the similarity between the \(t\)-th frame feature \(\mathbf{s}_{t,k,j}\) and prototype \(c_m\). 
Here, \(F_1\) and \(F_2\) are projection heads, \(\tau_1\) and \(\tau_2\) are temperature parameters, and \(\mathcal{I}\) indexes all frames in the training set.

\paragraph{Structure-Trajectory Prompted Reconstruction Loss (\( \mathcal{L}_{\text{STPR}} \)).}
To leverage spatial and temporal contexts, SGT introduces a \textbf{Structure-Trajectory Prompted Reconstruction Loss} that reconstructs masked joints and motion trajectories using unmasked features. 
The final loss is a weighted sum of two components:
\begin{equation}
\mathcal{L}_{\text{STPR}} = \beta \mathcal{L}_{\text{STPR}}^{\text{st}} + (1 - \beta) \mathcal{L}_{\text{STPR}}^{\text{tr}},
\end{equation}
where the structure-prompted and trajectory-prompted losses are respectively defined as:
\begin{align}
\mathcal{L}_{\text{STPR}}^{\text{st}} &= \frac{1}{N} 
\sum_{i=1}^{N} \big\| \hat{X}_i - X_i \big\|_1, \\
\mathcal{L}_{\text{STPR}}^{\text{tr}} &= \frac{1}{N} 
\sum_{i=1}^{N} \big\| \tilde{X}_i - X_i \big\|_1.
\end{align}
Here, \(X_i\) is the ground-truth skeleton, while \(\hat{X}_i\) and \(\tilde{X}_i\) denote reconstructions guided by structural and trajectory prompts, respectively.

\paragraph{Final Training Objective.}
The skeleton encoder is optimized by jointly minimizing the contrastive and reconstruction losses:
\begin{equation}
\mathcal{L} = \lambda \mathcal{L}_{\text{GPC}} + (1 - \lambda) \mathcal{L}_{\text{STPR}},
\end{equation}
where \(\lambda\) controls the trade-off between prototype contrastive learning and prompt-based reconstruction.

In summary, SGT extracts frame-level skeleton graph features through full-relation attention and refines them with \(\mathcal{L}_{\text{GPC}}\) and \(\mathcal{L}_{\text{STPR}}\), resulting in discriminative and semantically rich representations that are highly effective for skeleton-based person re-identification.

\begin{figure*}[t]
	\centering
	\includegraphics[width=1.0\textwidth]{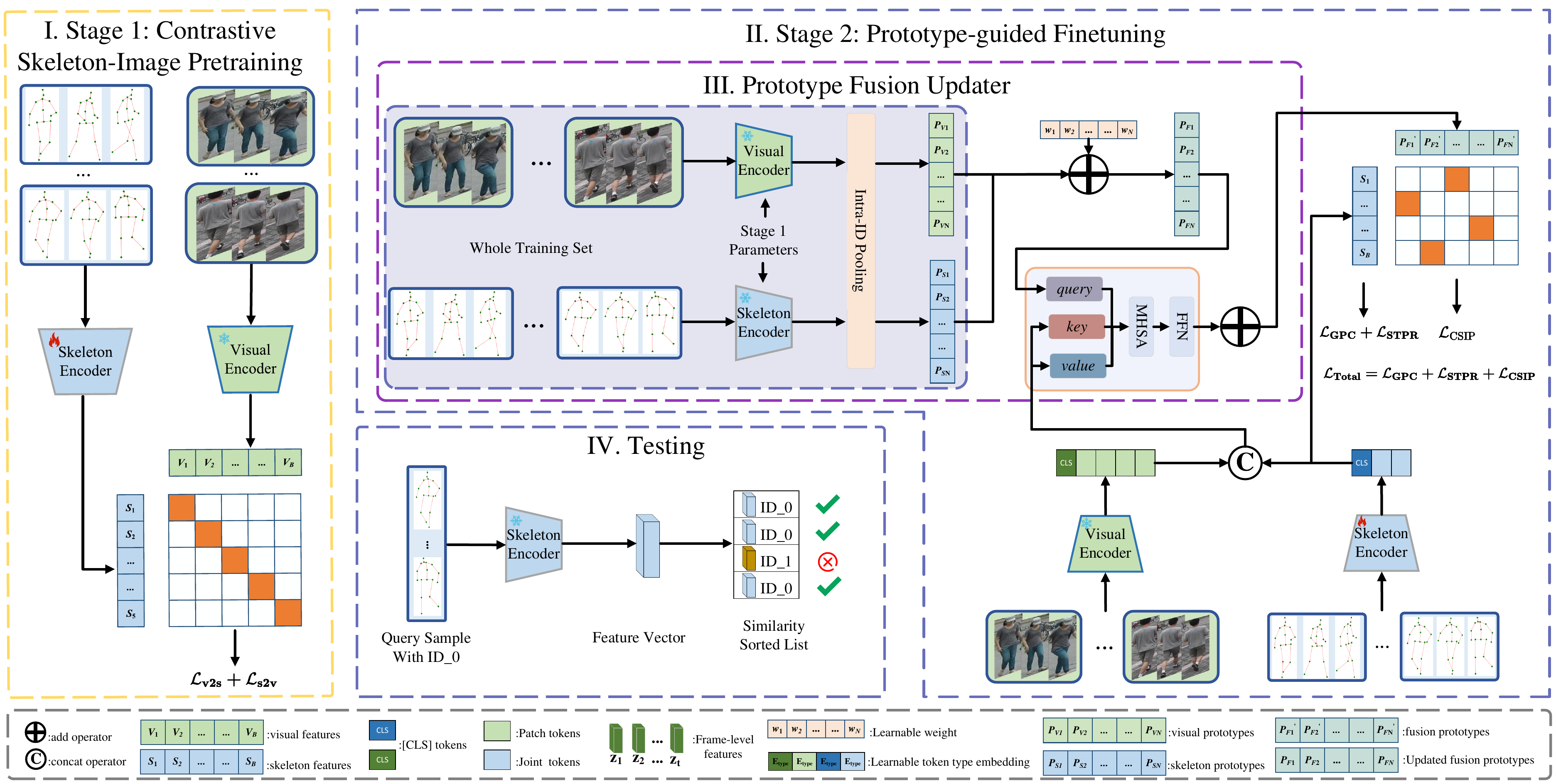}
	\caption{Overview of the proposed CSIP-ReID framework for skeleton-based person ReID. \textbf{(I) Stage1: Contrastive Skeleton-Image Pretraining}, where the skeleton encoder is fixed and the visual encoder is trained to align skeleton and visual features via supervised contrastive learning; \textbf{(II) Stage2: Prototype-guided Finetuning}, where the skeleton encoder and Prototype Fusion Updater (PFU) are optimized to refine modality-specific prototypes and enhance feature alignment; and \textbf{(III) Prototype Fusion Updater}, which constructs and dynamically updates visual and skeleton prototypes during training. \textbf{(IV) Testing} involves extracting skeleton features with the encoder trained in Stage 2 and ranking gallery samples based on feature similarity.}

	\label{fig:modified_model_structure}
\end{figure*}

\section{Training Strategy}

Take video-based ReID for example, our training strategy consists of two stages: Contrastive Skeleton-Image Pretraining and Prototype-guided Finetuning. 
In Stage 1, we load pretrained weights from CLIP, freeze the visual encoder and \textbf{\textit{optimize only the skeleton encoder}} to align the two modalities via supervised contrastive learning. The training objective is:
\begin{equation}
\mathcal{L}_{\text{stage } 1} = \mathcal{L}_{\text{v2s}} + \mathcal{L}_{\text{s2v}}.
\end{equation}

In Stage 2, we \textbf{\textit{jointly optimize the visual encoder, Prototype Fusion Updater (PFU), and Skeleton-Guided Temporal Modeling (SGTM)}}. Specifically, we employ the cross-entropy loss $\mathcal{L}_{\text{CE}}$, triplet loss $\mathcal{L}_{\text{Triplet}}$, prototype-guided supervision loss $\mathcal{L}_{\text{CSIP}}$ from PFU, and frame-level supervision loss $\mathcal{L}_{\text{Frame}}$ from SGTM. Two hyperparameters $\lambda_1$ and $\lambda_2$ control the contribution of the last two terms:

\begin{equation}
\mathcal{L}_{\text{stage } 2} = \mathcal{L}_{\text{CE}} + \mathcal{L}_{\text{Triplet}} + \lambda_1 \mathcal{L}_{\text{CSIP}} + \lambda_2 \mathcal{L}_{\text{Frame}}.
\end{equation}

The whole training Procedure of the proposed CSIP-ReID for video-based ReID is summarized in Algorithm~\ref{alg:training}.

\begin{algorithm}[t]
\caption{Training Procedure of CSIP-ReID}
\label{alg:training}
\begin{algorithmic}[1]
\REQUIRE Paired skeleton–image training dataset $\mathcal{D}=\{(\mathbf{I}_i, \mathbf{S}_i, y_i)\}_{i=1}^{N}$; 
hyperparameters $\lambda_1$, $\lambda_2$
\ENSURE Trained CSIP-ReID model with \textbf{\textit{visual encoder}} $\mathcal{V}(\cdot)$, \textbf{\textit{skeleton encoder}} $\mathcal{S}(\cdot)$, PFU, and SGTM

\vspace{2pt}
\STATE \textbf{Stage 1: Contrastive Skeleton-Image Pretraining}
\STATE Load pretrained CLIP weights; freeze $\mathcal{V}(\cdot)$
\STATE Initialize $\mathcal{S}(\cdot)$
\FOR{each mini-batch $(\mathbf{I}, \mathbf{S}, y)$ sampled as paired skeleton–image sequences from $\mathcal{D}$}
    \STATE Extract $\mathbf{v}_t = \mathcal{V}(\mathbf{I})$, $\mathbf{s}_t = \mathcal{S}(\mathbf{S})$
    \STATE Compute contrastive loss $\mathcal{L}_{\text{stage1}} = \mathcal{L}_{\text{v2s}} + \mathcal{L}_{\text{s2v}}$
    \STATE Update only $\mathcal{S}(\cdot)$ using gradient descent
\ENDFOR

\vspace{2pt}
\STATE \textbf{Stage 2: Prototype-guided Finetuning}
\STATE Unfreeze $\mathcal{V}(\cdot)$; initialize PFU and SGTM
\FOR{each mini-batch $(\mathbf{I}, \mathbf{S}, y)$ sampled using PK sampling from $\mathcal{D}$}
    \STATE Extract $\mathbf{v}_t = \mathcal{V}(\mathbf{I})$, $\mathbf{s}_t = \mathcal{S}(\mathbf{S})$
    \STATE construct fusion prototypes through PFU
    \STATE Update fusion prototypes through PFU
    \STATE Perform temporal modeling using SGTM
    \STATE Compute losses: classification loss $\mathcal{L}_{\text{CE}}$, triplet loss $\mathcal{L}_{\text{Tri}}$, CSIP loss $\mathcal{L}_{\text{CSIP}}$, and frame-level loss $\mathcal{L}_{\text{Frame}}$
    \STATE Compute total loss: 
    \STATE \quad $\mathcal{L}_{\text{stage2}} = \mathcal{L}_{\text{CE}} + \mathcal{L}_{\text{Tri}} + \lambda_1 \mathcal{L}_{\text{CSIP}} + \lambda_2 \mathcal{L}_{\text{Frame}}$
    \STATE Update $\mathcal{V}(\cdot)$, PFU, and SGTM via gradient descent
\ENDFOR
\STATE \textbf{return} Trained CSIP-ReID model
\end{algorithmic}
\end{algorithm}

\section{Additional Experiments}

\subsection{Details of ReID datasets}

We evaluate CSIP-ReID on three video-based person ReID benchmarks: \textbf{\textit{MARS}}~\cite{zheng2016mars}, \textbf{\textit{LS-VID}}~\cite{li2019global}, and \textbf{\textit{iLIDS-VID}}~\cite{wang2014person}. Furthermore, to assess its generalization capability, we also conduct experiments on two skeleton-based ReID datasets, \textbf{\textit{BIWI}}~\cite{munaro2014one} and \textbf{\textit{IAS}}~\cite{munaro2014feature}.

\textbf{\textit{MARS}}~\cite{zheng2016mars} is a large-scale benchmark for video-based person re-identification, captured by six cameras in a campus environment. 
The dataset contains 20,715 video sequences, including 17,503 tracklets from 1,261 identities and an additional 3,248 distractor tracklets. 
Among them, 625 identities are designated for training, while the remaining 636 identities are reserved for testing. 
Each tracklet is generated by detecting pedestrians using the DPM detector~\cite{4587597} and tracking them with the GMMCP tracker~\cite{7299036}, resulting in numerous sequences with misalignments, occlusions, and low-quality frames, which makes MARS a challenging dataset for video-based ReID research.

\textbf{\textit{LS-VID}}~\cite{li2019global} is a large-scale and challenging video-based person re-identification benchmark. It contains 14,943 video tracklets from 3,772 identities, resulting in approximately 2.98 million frames. The dataset is captured by 15 cameras, including both indoor and outdoor scenes, and the pedestrian bounding boxes are generated using the Faster R-CNN~\cite{ren2015faster} detector, ensuring higher detection quality. LS-VID provides rich multi-camera viewpoints and lighting variations, making it a valuable benchmark for advancing video-based ReID research.

\textbf{\textit{iLIDS-VID}}~\cite{wang2014person} is a small-scale video-based person re-identification dataset captured by two non-overlapping cameras in an airport arrival hall. It consists of 600 video sequences from 300 identities, with each identity appearing in both camera views. The dataset is particularly challenging due to frequent background clutter, severe occlusions, and significant illumination variations, making it a widely used benchmark for evaluating video-based ReID methods under complex real-world conditions.

\textbf{\textit{IAS-Lab RGBD-ID}}~\cite{munaro2014feature} is a skeleton-based person re-identification dataset acquired with RGB-D cameras in an indoor lab environment. It consists of 11 identities, each recorded in three sequences: a training sequence, a testing sequence A with different clothes, and a testing sequence B captured in a different room with the same clothes as the training set. The dataset provides synchronized RGB and depth images, user segmentation maps, and 3D skeleton joints estimated by OpenNI and NiTE middleware at 30 fps. Only frames where all joints are reliably tracked are retained, making IAS-Lab RGBD-ID a valuable benchmark for evaluating both short-term and long-term skeleton-based ReID methods.

\textbf{\textit{BIWI RGBD-ID}}~\cite{munaro2014one} is a skeleton-based person re-identification dataset collected with a Microsoft Kinect for Windows, aimed at evaluating long-term ReID. It consists of 50 subjects with 50 training sequences and 56 testing sequences. Training videos capture individuals performing rotations, head movements, and walking routines in front of the camera. For 28 subjects, testing videos were recorded on different days and in different locations, often with different clothing, to increase appearance variability. Each testing subject has a Still sequence (standing or slight motion) and a Walking sequence (two frontal and two diagonal walks). The dataset provides synchronized high-resolution RGB (1280×960) and depth images, segmentation maps, and 3D skeleton data (tracked by the Microsoft Kinect SDK) at 10 fps, making it a valuable benchmark for skeleton-based ReID under realistic cross-day and cross-clothing variations.

\subsection{Experiment settings}

Our model is implemented on the PyTorch platform and trained on a single NVIDIA Tesla L20 GPU with 48 GB memory. The visual encoder is based on ViT-B/16 from CLIP~\cite{radford2021learning}, while the skeleton encoder adopts the Skeleton Graph Transformer from TranSG~\cite{rao2023transg}. Each tracklet is sampled with 8 frames, resized to 256$\times$128, and augmented following the strategy in TF-CLIP~\cite{tfclip_2024}. 
Stage 1 is trained for 120 epochs with a batch size of 64, and Stage 2 is trained for 80 epochs using PK sampling~\cite{hermans2017defense} with 4 identities and 4 tracklets per identity. Following TF-CLIP~\cite{tfclip_2024}, we employ the Adam optimizer~\cite{kingma2014adam}, warming up the model for the first 10 epochs with a linearly increasing learning rate from $5 \times 10^{-7}$ to $5 \times 10^{-6}$. The learning rate is then decayed by a factor of 10 at the 30th, 50th, and 70th epochs. The loss balancing hyperparameters $\lambda_1$ and $\lambda_2$ are set to 1.0 and 1.3, respectively. Euclidean distance is used as the ranking metric during evaluation.
For the skeleton encoder, we follow the settings in TranSG~\cite{rao2023transg} and adopt the Human3.6M skeleton connectivity, which defines 17 joints as illustrated in Fig.~\ref{fig:modified_model_structure}.

\subsection{Model structure for skeleton-based ReID}

As illustrated in Fig.~\ref{fig:modified_model_structure}, the CSIP-ReID architecture for skeleton-based ReID is slightly different from the one used for video-based ReID. 

In \textbf{\textit{Stage 1: Contrastive Skeleton-Image Pretraining}}, we fix the skeleton encoder and optimize only the visual encoder. The two modalities are aligned through supervised contrastive learning, enabling the encoders to produce well-aligned features that combine appearance-rich visual cues with motion-aware skeleton representations. It is important to note that the fixed skeleton encoder is initialized with weights pre-trained directly on the skeleton-based ReID datasets.  

In \textbf{\textit{Stage 2: Prototype-Guided Finetuning}}, we optimize only the skeleton encoder together with the Prototype Fusion Updater (PFU). The Skeleton-Guided Temporal Modeling (SGTM) module is removed in this stage because it was originally introduced to compensate for the lack of temporal modeling capability in the Vision Transformer used as the visual encoder. In contrast, for the skeleton-based ReID task, the skeleton encoder is implemented as a skeleton graph transformer, which inherently exhibits strong temporal modeling capacity, making the SGTM module unnecessary in this setting.

During the \textbf{\textit{testing stage}}, each query skeleton sequence is first processed by the skeleton encoder, which loads the weights obtained at the end of Stage 2, to extract its discriminative feature representation. The resulting feature vector is then compared against all gallery features to compute pairwise similarities. Finally, the gallery samples are ranked according to their similarity scores, forming a similarity-sorted list. The retrieval is considered correct if samples with the same identity as the query appear at the top of the list, while the presence of mismatched identities in higher ranks leads to retrieval errors.

\subsection{Additional Performance Comparison on Video-based and Skeleton-based ReID}

Tab.~\ref{tab:full_video_based} and Tab.~\ref{tab:full_skeleton_based} report the complete experimental results comparing our CSIP-ReID with state-of-the-art methods on both video-based and skeleton-based person ReID benchmarks. The results demonstrate that CSIP-ReID consistently achieves superior performance across all datasets.

For \textbf{video-based ReID} (Tab.~\ref{tab:full_video_based}), CSIP-ReID surpasses all competing approaches on MARS, LS-VID, and iLIDS-VID. On the MARS dataset, our method achieves the highest mAP of \textbf{90.4\%} and Rank-1 accuracy of \textbf{94.2\%}, outperforming recent strong baselines such as CLIMB-ReID and TF-CLIP. On LS-VID, CSIP-ReID also sets a new state-of-the-art with \textbf{85.0\%} mAP and \textbf{92.5\%} Rank-1 accuracy, significantly improving over the previous best result of 85.0\% mAP and 91.3\% Rank-1 achieved by CLIMB-ReID. Furthermore, on the challenging iLIDS-VID dataset, CSIP-ReID obtains the best Rank-1 accuracy of \textbf{97.2\%} and Rank-5 accuracy of \textbf{98.2\%}, demonstrating its robustness in complex surveillance scenarios.

For \textbf{skeleton-based ReID} (Tab.~\ref{tab:full_skeleton_based}), CSIP-ReID outperforms all hand-crafted (H), sequence learning (S), and graph-based (G) methods by a large margin on BIWI and IAS datasets. Specifically, on BIWI-S, our approach achieves \textbf{34.5\%} mAP and \textbf{86.1\%} Rank-1, substantially exceeding the previous best of 32.1\% mAP and 79.3\% Rank-1 by MoCos. Similar improvements are observed on BIWI-W, where CSIP-ReID obtains \textbf{33.8\%} mAP and \textbf{59.1\%} Rank-1, outperforming all competitors. On IAS-A and IAS-B, our method consistently achieves the highest Rank-1 accuracies of \textbf{76.9\%} and \textbf{63.3\%}, respectively, indicating its strong capability to model both short-term and long-term skeleton-based identity cues.

Overall, these results clearly demonstrate the effectiveness of our proposed CSIP-ReID, which not only sets new state-of-the-art performance on multiple video-based benchmarks but also significantly advances the state of the art on skeleton-based ReID tasks.

\section{Retrieval result visualization}

\begin{figure*}[t]
	\centering
	\includegraphics[width=1.0\textwidth]{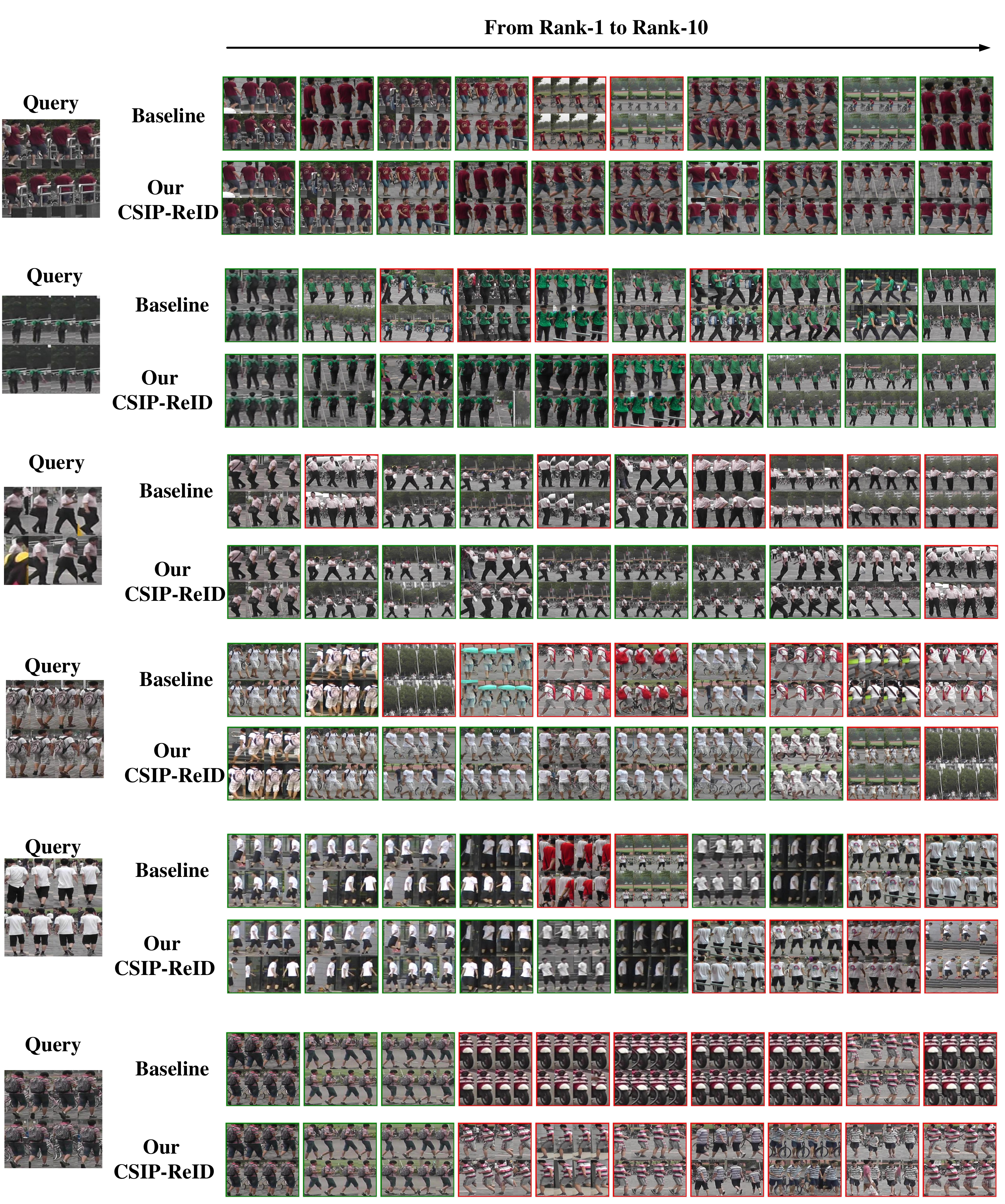}
	\caption{\textbf{Visualization of pedestrian retrieval results} on the MARS dataset, comparing the baseline with our proposed CSIP-ReID. Each image represents a sequence sample composed of 8 consecutive frames. For each query, the top-10 retrieved sequences are shown from left to right. \textbf{Green} bounding boxes indicate correct matches, while \textbf{red} bounding boxes denote incorrect predictions. The results highlight that CSIP-ReID achieves more accurate retrieval and better ranking performance than the baseline, particularly under challenging conditions.}

	\label{fig:retrieval}
\end{figure*}

We visualize the retrieval results of the baseline and CSIP-ReID on the MARS dataset in Fig.~\ref{fig:retrieval}. \textbf{Green} boxes indicate correct matches, while \textbf{red} boxes denote errors. The examples show that CSIP-ReID consistently outperforms the baseline, delivering more accurate retrieval and superior ranking, especially in challenging scenarios.

\textbf{\textit{In the first example}} under a normal scenario, the baseline method makes two errors at the 4th and 5th ranks, while CSIP-ReID successfully retrieves all correct sequences, demonstrating its robustness.
\textbf{\textit{The second example}} corresponds to the green-clothed individual also illustrated in the t-SNE visualization in the main text. In the MARS test set, nearly 20 different identities wear similar green outfits. The baseline produces four errors within the top-10 results, including three incorrect matches in the top-5, while CSIP-ReID makes only one error in the top-10, demonstrating its stronger discriminative capability in scenarios with high appearance similarity.  

\textbf{\textit{In the third example}}, the misidentified samples by the baseline exhibit appearance features almost indistinguishable from the target, making them difficult to differentiate even by human inspection. However, CSIP-ReID correctly identifies the first nine samples, \textbf{\textit{likely benefiting from the skeleton features that capture fine-grained motion cues}}, enabling it to distinguish between individuals with highly similar appearances.  
\textbf{\textit{In the fourth example}}, the baseline focuses heavily on the presence of a backpack in the query, retrieving mostly backpack-wearing pedestrians, some of which are clearly incorrect. In contrast, CSIP-ReID does not overly rely on the backpack cue, \textbf{\textit{retrieving correct matches even when the target individual is not carrying a backpack}}, again likely due to the model’s ability to exploit subtle motion cues.  

\textbf{\textit{The fifth example}} shows that although both methods retrieve the same top-10 samples, CSIP-ReID produces a more accurate ranking, demonstrating its superior feature ordering capability.  
Finally, \textbf{\textit{the sixth example}} presents an extreme case where the query sample is partially occluded by a backpack and only the back view is visible, providing very limited appearance information. The baseline fails significantly, focusing on visually similar motorbikes, whereas CSIP-ReID, although unable to perfectly identify the target, \textbf{\textit{directs its attention toward human regions}} thanks to the utilization of distilled skeleton information. This behavior is consistent with the Focus Region Analysis discussed in the main text.

\section{Discussion}
\subsection{Towards Large-Scale Skeleton-Image Pretraining}

Although CSIP-ReID achieves outstanding performance, there remains considerable room for improvement in the pretraining stage. Currently, pretraining is performed separately on each dataset, resulting in weights that are tailored to dataset-specific feature distributions. Compared with the large-scale pretraining of CLIP, the ReID datasets used in our work are relatively small, limiting the potential benefits of pretraining.  

To overcome this limitation, a promising direction is to jointly leverage multiple ReID datasets to construct a larger paired skeleton–image database. This approach would not only enlarge the training scale but also enable the pretrained weights to generalize across diverse feature distributions. Future work will explore multi-dataset pretraining to further enhance skeleton–image sequence alignment.  

Beyond ReID, contrastive skeleton–image pretraining has the potential for broader applications. Since it only requires paired skeleton–image data, it can be extended to other domains such as action recognition, gait recognition, and pedestrian tracking. By extracting skeletons from videos or images in these domains, we can build a large-scale cross-domain pretraining dataset. This strategy positions CSIP as a promising foundation model, with the capacity to benefit a wide range of vision tasks.

\subsection{Comparison with other ReID tasks}

The ReID field encompasses a variety of specialized sub-tasks, among which video-based ReID, addressed by our CSIP-ReID, is only one example. To provide readers with a broader understanding of the ReID domain, this section reviews several emerging ReID tasks that have gained attention in recent years, along with representative methods for each. We further analyze their respective strengths and weaknesses to highlight the unique challenges and research opportunities within these sub-domains.

\subsubsection{Video-based ReID.}
Video-based person re-identification seeks to learn robust identity representations by capturing discriminative spatio-temporal cues from video sequences. TF-CLIP~\cite{tfclip_2024} adopts a CLIP-style framework, replacing the text encoder with a visual memory module and introducing a temporal memory diffusion module to model temporal dependencies. In contrast, we propose Contrastive Skeleton-Image Pretraining (CSIP-ReID), which leverages prototype learning and the Learning Using Privileged Information (LUPI)~\cite{vapnik2009new} paradigm to distill rich motion cues from skeleton data and enhance visual feature learning.

\subsubsection{Skeleton-based ReID.} Skeleton-based person ReID focuses on the problem of matching and retrieving a certain person based on spatial and temporal representations of skeletal human body and gait. 
TranSG~\cite{rao2023transg} introduces a Skeleton Graph Transformer with prototype contrastive learning and structure-trajectory reconstruction to capture fine-grained spatio-temporal patterns for person re-identification but struggles with noisy or incomplete skeletons.
MoCos~\cite{rao2025motif} employs a motif-guided skeleton graph transformer with combinatorial prototype learning to enhance discriminative skeleton representations for person re-identification, but relies on well-defined motifs and random masking.

\subsubsection{VI ReID.}  Visible-Infrared ReID aims to retrieve images across visible and infrared modalities, suffering from the absence of critical information, e.g. color, in infrared modality.
TVI-LFM~\cite{2024TVI-LFM} leverages VLM- and LLM-generated descriptions to enrich infrared representations and improve cross-modal retrieval, but its performance is sensitive to the accuracy and richness of the generated texts.
USVI-ReID~\cite{shi2024learningcommonalitydivergencevariety} introduces a progressive contrastive learning framework with hard and dynamic prototypes that effectively captures commonality, divergence, and variety, but it relies on DBSCAN-generated pseudo-labels.

\subsubsection{T2I ReID.} Text-to-image person re-identification aims to match a person image with a given natural language description from a large gallery set. 
ICL~\cite{qin2025human} employs MLLM-driven interactive reasoning and text augmentation to enhance the accuracy and generalization of text-to-image person re-identification, but its performance is constrained by the quality of MLLM-generated descriptions.
Tan et al. ~\cite{tan2024harnessing} leverage MLLMs to generate diverse text descriptions and apply a similarity-based masking strategy to handle noisy annotations, achieving strong cross-dataset generalization in text-to-image person re-identification.

\subsubsection{Multimodal ReID.} Multimodal person ReID focuses on the problem of matching and retrieving a target individual by leveraging and aligning complementary information from multiple modalities such as RGB, infrared, sketches, and text descriptions.
AIO~\cite{li2024all} is a unified multimodal ReID framework that employs a frozen Transformer foundation model and a lightweight tokenizer to align RGB, IR, Sketch, and Text in a shared space, achieving strong zero-shot generalization but remaining limited by its reliance on synthetic data and restricted adaptability.
FlexiReID introduces an adaptive mixture-of-experts framework with cross-modal query fusion to achieve flexible person retrieval across arbitrary combinations of RGB, infrared, sketch, and text modalities.

\subsubsection{Occluded ReID} The task of occluded person ReID is to find the same person under different cameras while the target pedestrian is obscured. 
FPC~\cite{ye2024dynamic} introduces a token sparsification and multi-view feature consolidation framework that suppresses occlusion noise and restores missing features, but its performance relies on accurate neighbor retrieval and involves higher inference complexity.
SPT~\cite{tan2024occluded} introduces a saliency-guided patch transfer strategy that leverages real occlusion information and occlusion-aware IoU to generate high-quality occluded samples, thereby enhancing the robustness of ViT-based person ReID models under occlusion but with limited validation on diverse datasets and architectures.

% ----------------------------------------------------------------------------------------------------
\clearpage

\begin{table*}[]
    \centering
    \resizebox{1.00\textwidth}{!}{%
        \begin{tabular}{cccccccc}
\hline
\multirow{2}{*}{Methods} & \multirow{2}{*}{Source} & \multicolumn{2}{c}{MARS} & \multicolumn{2}{c}{LS-VID} & \multicolumn{2}{c}{iLIDS-VID} \\
                         &                         & mAP & Rank-1               & mAP & Rank-1                & Rank-1 & Rank-5                \\ \hline
STMP~\cite{liu2019spatial}         & AAAI19   & 72.7  & 84.4  & 39.1  & 56.8  & 84.3  & 96.8  \\
M3D~\cite{li2019multi}             & AAAI19   & 74.1  & 84.4  & 40.1  & 57.7  & 74.0  & 94.3  \\
GLTR~\cite{li2019global}           & ICCV19   & 78.5  & 87.0  & 44.3  & 63.1  & 86.0  & 98.0  \\
TCLNet~\cite{hou2020temporal}      & ECCV20   & 85.1  & 89.8  & 70.3  & 81.5  & 86.6  & -     \\
MGH~\cite{yan2020learning}         & CVPR20   & 85.8  & 90.0  & 61.8  & 79.6  & 85.6  & 97.1  \\
GRL~\cite{liu2021watching}         & CVPR21   & 84.8  & 91.0  & -     & -     & 90.4  & 98.3  \\
BiCnet-TKS~\cite{hou2021bicnet}    & CVPR21   & 86.0  & 90.2  & 75.1  & 84.6  & -     & -     \\
CTL~\cite{liu2021spatial}          & CVPR21   & 86.7  & 91.4  & -     & -     & 89.7  & 97.0  \\
STMN~\cite{eom2021video}           & ICCV21   & 84.5  & 90.5  & 69.2  & 82.1  & -     & -     \\
PSTA~\cite{wang2021pyramid}        & ICCV21   & 85.8  & 91.5  & -     & -     & 91.5  & 98.1  \\
DIL~\cite{he2021dense}             & ICCV21   & 87.0  & 90.8  & -     & -     & 92.0  & 98.0  \\
% STT~\cite{zhang2021spatiotemporal} & Arxiv21  & 86.3  & 88.7  & 78.0  & 87.5  & 87.5  & 95.0  \\
CAVIT~\cite{wu2022cavit}           & ECCV22   & 87.2  & 90.8  & 79.2  & 89.2  & 93.3  & 98.0  \\
SINet~\cite{bai2022salient}        & CVPR22   & 86.2  & 91.0  & 79.6  & 87.4  & 92.5  & -     \\
MFA~\cite{gu2022motion}            & TIP22    & 85.0  & 90.4  & 78.9  & 88.2  & 93.3  & 98.7  \\
DCCT~\cite{liu2023deeply}          & TNNLS23  & 87.5  & 92.3  & -     & -     & 91.7  & 98.6  \\
TMT~\cite{10534084}                & TITS24   & 85.8  & 91.2  & -     & -     & 91.3  & 98.6  \\
MS-STI~\cite{ran2024multiscale}    & TCSVT24  & 87.2  & 92.7  & 80.7  & 89.6  & -     & -  \\
TCVIT~\cite{wu2024temporal} & AAAI24 & 87.6 & 91.7 & 83.1 & 90.1 & 94.3 & \underline{99.3}  \\
TF-CLIP~\cite{tfclip_2024} & AAAI24 & 89.4 & 93.0 & 83.8 & 90.4 & 94.5  & 99.1  \\
TAE-ViT~\cite{wang2025learning} & ESWA25 & 86.7 & 90.2 & - & - & 93.3  & 98.7  \\
3DAPRL~\cite{jing20253d} & TCSVT25 & \underline{90.3} & 93.1 & - & - & 94.7  & 98.7  \\
CLIMB-ReID~\cite{Yu_Liu_Zhu_Wang_Zhang_Lu_2025} & AAAI25 & 89.7 & \underline{93.3} & \textbf{85.0} & \underline{91.3} & \underline{96.7} & \textbf{99.9} \\
\textbf{CSIP-ReID(Ours)} &                    & \textbf{90.4} & \textbf{94.2} & \underline{84.2} & \textbf{92.5} & \textbf{97.2} & 98.2 \\ 
\hline
\end{tabular}
        }
    \caption{Comparison with state-of-the-art methods on MARS, LS-VID and iLIDS-VID for video-based ReID. \textbf{Bold} numbers indicate the best performance, while \underline{underlined} numbers denote the second-best results.}
    \label{tab:full_video_based}
\end{table*}

% ----------------------------------------------------------------------------------------------------

\begin{table*}[t]
\centering
\scalebox{1.00}{
\renewcommand\arraystretch{1.3}{
\setlength{\tabcolsep}{1.2mm}{
\begin{tabular}{cl|rrrr|rrrr|rrrr|rrrr}
\hline
\multicolumn{2}{c|}{\multirow{2}{*}{\textbf{Methods}}} & \multicolumn{4}{c|}{\textbf{BIWI-S}} & \multicolumn{4}{c|}{\textbf{BIWI-W}} & \multicolumn{4}{c|}{\textbf{IAS-A}} & \multicolumn{4}{c}{\textbf{IAS-B}} \\ \cline{3-18} 
\multicolumn{2}{l|}{} & \textbf{mAP} & \textbf{R$_{1}$} & \textbf{R$_{5}$} & \textbf{R$_{10}$} & \textbf{mAP} & \textbf{R$_{1}$} & \textbf{R$_{5}$} & \textbf{R$_{10}$} & \textbf{mAP} & \textbf{R$_{1}$} & \textbf{R$_{5}$} & \textbf{R$_{10}$} & \textbf{mAP} & \textbf{R$_{1}$} & \textbf{R$_{5}$} & \textbf{R$_{10}$} \\ \hline
\multicolumn{1}{l|}{\multirow{3}{*}{\textbf{H.}}} & ${D_{\text{PG}}}$ \shortcite{liao2020model} & 6.7 & 18.5 & 45.4 & 63.8 & 8.7 & 6.5 & 15.5 & 20.3 & 11.0 & 16.4 & 39.5 & 53.4 & 10.6 & 16.0 & 41.2 & 57.3 \\
\multicolumn{1}{l|}{} & ${D_{13}}$ \shortcite{munaro2014one} & 13.1 & 28.3 & 53.1 & 65.9 & 17.2 & 14.2 & 20.6 & 23.7 & 24.5 & 40.0 & 58.7 & 67.6 & 23.7 & 43.7 & 68.6 & 76.7 \\
\multicolumn{1}{l|}{} & ${D_{16}}$ \shortcite{pala2019enhanced} & 16.7 & 32.6 & 55.7 & 68.3 & 18.8 & 17.0 & 25.3 & 29.6 & 25.2 & 42.7 & 62.9 & 70.7 & 24.5 & 44.5 & 69.1 & 80.2 \\ \hline
\multicolumn{1}{l|}{\multirow{5}{*}{\textbf{S.}}} & PoseGait \shortcite{liao2020model} & 9.9 & 14.0 & 40.7 & 56.7 & 11.1 & 8.8 & 23.0 & 31.2 & 17.5 & 28.4 & 55.7 & 69.2 & 20.8 & 28.9 & 51.6 & 62.9 \\
\multicolumn{1}{l|}{} & AGE \shortcite{rao2020self} & 8.9 & 25.1 & 43.1 & 61.6 & 12.6 & 11.7 & 21.4 & 27.3 & 13.4 & 31.1 & 54.8 & 67.4 & 12.8 & 31.1 & 52.3 & 64.2 \\
\multicolumn{1}{l|}{} & SGELA \shortcite{rao2021self} & 15.1 & 25.8 & 51.8 & 64.4 & 19.0 & 11.7 & 14.0 & 14.7 & 13.2 & 16.7 & 30.2 & 44.0 & 14.0 & 22.2 & 40.8 & 50.2 \\
\multicolumn{1}{l|}{} & SimMC \shortcite{rao2022simmc} & 12.3 & 41.7 & 66.6 & 76.8 & 19.9 & 24.5 & 36.7 & 44.5 & 18.7 & 44.8 & 65.3 & 72.9 & 22.9 & 46.3 & 68.1 & 77.0 \\
\multicolumn{1}{l|}{} & Hi-MPC \shortcite{rao2024hierarchical} & 17.4 & 47.5 & 70.3 & 78.6 & 22.6 & 27.3 & 40.3 & 48.8 & 23.2 & 45.6 & 67.3 & 75.4 & 25.3 & 48.2 & 70.2 & 77.8 \\ \hline

\multicolumn{1}{l|}{\multirow{6}{*}{\textbf{G.}}} & MG-SCR \shortcite{rao2021multi} & 7.6 & 20.1 & 46.9 & 64.1 & 11.9 & 10.8 & 20.3 & 29.4 & 14.1 & 36.4 & 59.6 & 69.5 & 12.9 & 32.4 & 56.5 & 69.4 \\
\multicolumn{1}{l|}{} & SM-SGE \shortcite{rao2021sm} & 10.1 & 31.3 & 56.3 & 69.1 & 15.2 & 13.2 & 25.8 & 33.5 & 13.6 & 34.0 & 60.5 & 71.6 & 13.3 & 38.9 & 64.1 & 75.8 \\
\multicolumn{1}{l|}{} & SPC-MGR \shortcite{rao2022skeleton} & 16.0 & 34.1 & 57.3 & 69.8 & 19.4 & 18.9 & 31.5 & 40.5 & 24.2 & 41.9 & 66.3 & 75.6 & 24.1 & 43.3 & 68.4 & 79.4 \\
\multicolumn{1}{l|}{} & ST-GCN \shortcite{yan2018spatial} & 28.5 & 61.6 & 78.2 & 89.5 & 28.2 & 32.9 & 47.6 & 54.8 & 34.0 & 41.6 & 60.6 & 68.2 & 28.1 & 49.1 & 68.1 & 76.3 \\
\multicolumn{1}{l|}{} & TranSG \shortcite{rao2023transg} & 30.1 & \underline{68.7} & \underline{86.5} & 91.8 & 26.9 & 32.7 & 44.9 & 52.2 & 32.8 & 49.2 & 68.5 & 76.2 & 39.4 & 59.1 & 77.0 & 87.0 \\
\multicolumn{1}{l|}{} & MoCos \shortcite{rao2025motif} & \underline{32.1} & \textbf{72.0} & \textbf{89.5} & \textbf{93.0} & \underline{30.5} & \underline{36.0} & \underline{49.2} & \underline{57.0} & \underline{35.8} & \underline{51.9} & \underline{69.4} & \underline{77.5} & \underline{45.5} & \underline{61.5} & \underline{79.1} & \underline{87.8} \\

\multicolumn{1}{l|}{} & \textbf{CSIP-ReID (Ours)} & \textbf{34.5} & 68.6 & 86.1 & \underline{92.3} & \textbf{33.8} & \textbf{36.9} & \textbf{51.3} & \textbf{59.1} & \textbf{48.1} & \textbf{53.6} & \textbf{76.9} & \textbf{82.2} & \textbf{50.7} & \textbf{63.3} & \textbf{83.1} & \textbf{88.7}

\\ \hline

\end{tabular}
}
}
}
\caption{Comparison with state-of-the-art methods on BIWI and IAS datasets for skeleton-based ReID. Methods are categorized into \textbf{H}: hand-crafted approaches, \textbf{S}: sequence representation learning methods, and \textbf{G}: graph-based methods. \textbf{Bold} numbers indicate the best performance, while \underline{underlined} numbers denote the second-best results.}

\label{tab:full_skeleton_based}
\end{table*}

% ----------------------------------------------------------------------------------------------------

\clearpage

\end{document}